\pdfoutput=1

\documentclass[11pt]{article}

\usepackage[preprint]{acl}

\usepackage{times}
\usepackage{latexsym}

\usepackage[T1]{fontenc}

\usepackage[utf8]{inputenc}

\usepackage{microtype}

\usepackage{inconsolata}

\usepackage{graphicx}
\usepackage{adjustbox}
\usepackage{multirow}
\usepackage{caption}
\usepackage{array}
\usepackage{makecell}
\usepackage{tablefootnote}  

\usepackage{float}
\usepackage{placeins}
\usepackage{booktabs}
\usepackage{xspace}
\usepackage{comment}
\usepackage{graphicx}
\usepackage{amssymb}
\usepackage{pifont}

\usepackage{listings}

\usepackage{kotex}
\usepackage{arydshln}

\newcommand{\modelName}{\textsc{LangAlign}\xspace}
\newcommand{\revModelNM}{\textsc{Rev-LangAlign}\xspace}

\title{\modelName: Enhancing Non-English Language Models via Cross-Lingual Embedding Alignment}

\author{
    Jong Myoung Kim$^{1,2}$ \hspace{0.3cm}
    Young-Jun Lee$^{2}$ \hspace{0.3cm}
    Ho-Jin Choi$^{2}$ \hspace{0.3cm}
    Sangkeun Jung$^{3}$ \hspace{0.3cm}
    \\
    $^1$SK-telecom \hspace{0.3cm}
    $^2$School of Computing, KAIST \hspace{0.3cm} \\
    $^3$The Division of Computer Convergence, Chugnam National University\\
    \href{mailto:jmkim71@sk.com}{\tt gray.apl@gmail.com} \hspace{0.3cm} 
    \{\href{mailto:yj2961@kaist.ac.kr}{\tt yj2961},\href{mailto:hojinc@kaist.ac.kr}{\tt hojinc}\}{\tt @kaist.ac.kr} \hspace{0.3cm} 
    \href{mailto:hugman@cnu.ac.kr}{\tt hugman@cnu.ac.kr}
}

\begin{document}
\maketitle
\begin{abstract}
While Large Language Models have gained attention, many service developers still rely on embedding-based models due to practical constraints. In such cases, the quality of fine-tuning data directly impacts performance, and English datasets are often used as seed data for training non-English models.
In this study, we propose \modelName, which enhances target language processing by aligning English embedding vectors with those of the target language at the interface between the language model and the task header.
Experiments on Korean, Japanese, and Chinese demonstrate that \modelName significantly improves performance across all three languages. Additionally, we show that \modelName can be applied in reverse to convert target language data into a format that an English-based model can process.

\end{abstract}

\section{Introduction}
The rapid advancement of large language models (LLMs) has significantly improved AI services. However, due to high computational costs, latency, and limited controllability, embedding-based LMs remain essential for certain tasks. These models require effective tuning and large-scale domain-specific data—resources that are relatively scarce for the non-English language. This study addresses this limitation by leveraging English data.

We focus on a chatbot system that extracts user profiles from conversations and uses them to generate responses. This involves detecting and managing conflicts between existing profiles and newly extracted ones. To address this, we frame the task as a Natural Language Inference (NLI) problem using RoBERTa, while employing a larger LLM for the more complex task of profile extraction.

To minimize costs before full-scale dataset construction, we first experimented with English NLI data to assess its feasibility. English datasets offer advantages in quality, quantity, and diversity, making them a cost-effective alternative~\citep{wang2022super, achiam2023gpt, bubeck2023sparks, joshi2020state}. However, applying English data to Korean tasks typically requires translation into Korean, which improves performance~\citep{gaim2023question} but incurs significant costs. To address this, we propose Language Aligner (\modelName).

\modelName improves training efficiency by aligning English data embeddings with target-language embeddings between the LM and task head. This enables: (1) effective use of rich English datasets,
(2) retention of performance benefits from translated data, and
(3) reduction of data processing costs.

We evaluate \modelName on NLI and Sentiment Analysis (SA) tasks across Korean, Japanese, and Chinese. Experimental results show that \modelName outperforms English-only models and achieves performance comparable to models trained on target-language data, demonstrating its cost-effectiveness.

We also explore a reverse application of \modelName for transfer inference, transforming target-language inputs into English embedding space. This enables efficient inference using English-based models.

Our main contributions are as follows: 
\begin{itemize} 
    \item We propose \modelName and a training strategy to efficiently learn from foreign-language data.
    \item We introduce a reverse application of \modelName for transfer inference from non-English to English embedding space.
    \item We demonstrate that \modelName is more cost-effective than conventional dataset construction, while providing a strong performance benchmark. 
\end{itemize}

\section{Related Works}

Multilingual pre-trained LMs such as mBERT~\citep{devlin2018bert}, mT5~\citep{xue2020mt5}, and XLM-R~\citep{conneau2019unsupervised} have become standard backbones for fine-tuning in non-English NLP tasks.
Their performance, however, is known to vary significantly depending on the language of the training data~\citep{schuster2018cross, gaim2023question}, with a clear gap between native-language, translated, and unrelated-language data.

To reduce the cost of building native-language datasets, we propose a lightweight layer inserted between the LM and the task head to mitigate cross-lingual representation gaps at the embedding level.

Previous work~\citep{tan2022multilingual} has aligned multilingual sentence embeddings by enforcing similarity between translation pairs, typically through LM-level training.
In contrast, our method improves downstream task performance by inserting a small adaptation layer without modifying the backbone encoder.

Other approaches have leveraged multilingual knowledge bases~\citep{jiang2024pre} or enforced cross-lingual consistency during pretraining~\citep{gao2023learning}.
However, these rely on large-scale resources like parallel corpora or cross-lingual links, which may be impractical in low-resource settings.

Inspired by parameter-efficient adaptation methods~\citep{houlsby2019parameter, moosavi-etal-2022-adaptable, hu2021lora}, our approach introduces a simple yet effective bridging layer that enhances cross-lingual transferability without requiring additional large-scale data or retraining.

\section{\modelName}
\subsection{Language Aligner}
\begin{equation}
  \label{eq:langalign}
  \mathcal{A}(e(d_e)) \approx e(\mathcal{T}(d_e))
\end{equation}

Equation~\ref{eq:langalign} represents the role of \modelName. 
Here, \( d_e \), \( \mathcal{T}(d_e) \), \( e(d_e) \), \( e(\mathcal{T}(d_e)) \), and \( \mathcal{A}(e(d_e)) \) denote English data, English data translated into the target language, the embedding of English data, the embedding of translated data, and the embedding of English data processed through \modelName, respectively. \modelName is a layer that learns to convert the embedding of English data into the embedding of translated data.

Data processed through a LM is represented in a semantic space. We hypothesized that if we can transform \( e(d_e) \) to \( e(\mathcal{T}(d_e)) \), we can achieve transfer learning efficiency comparable to training on translated data. We designed two types of \modelName, with detailed descriptions of each layer provided in the appendix~\ref{ap:modelDetail}.

\paragraph{Fully Connected \modelName (FC)}
We hypothesized that transforming \( e(d_e) \) to \( e(\mathcal{T}(d_e)) \) could be represented using weights in a fully connected (FC) layer. Given the size of the embeddings output by the LM and cost considerations, we designed a simple FC structure. This \modelName consists of six layers, with ReLU activation and dropout applied between layers.

\paragraph{Auto Encoder \modelName (AE)}
We also explored the possibility of \modelName functioning as a translation mechanism, leading us to employ an Auto Encoder (AE) structure, which is well-suited for such tasks. \modelName AE includes an encoder and a decoder, each with three FC layers, where the layer dimensions are progressively halved or doubled. ReLU activations are applied between the layers.

\begin{figure*}[h]
    \centering
    \includegraphics[width=0.8\linewidth]{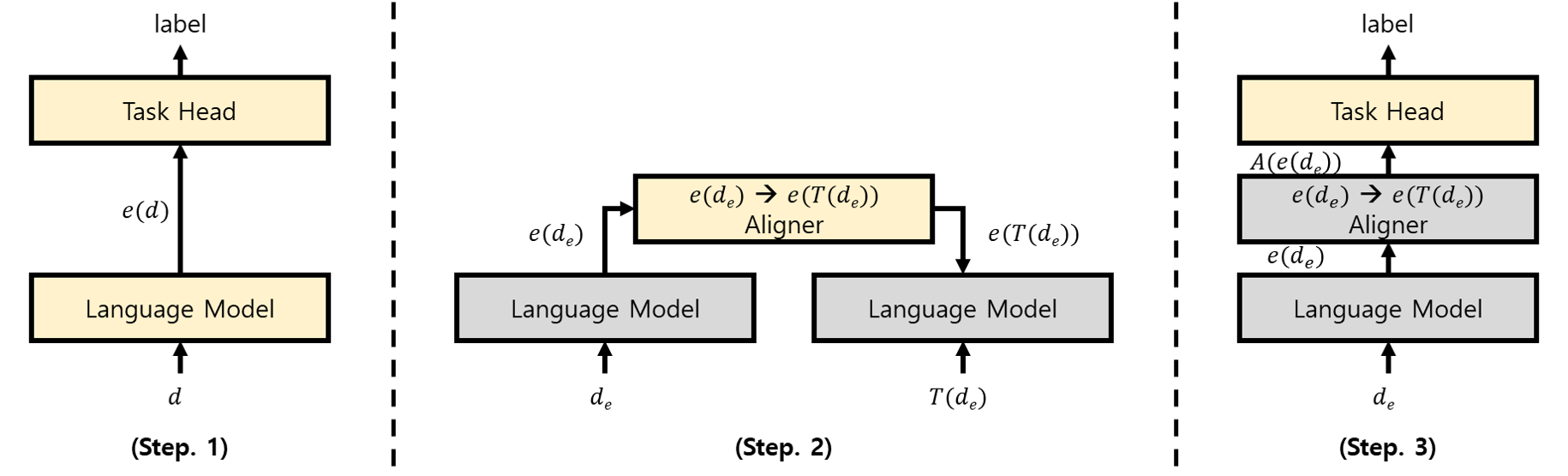} 
    \caption{Task fine-tuning sequence using \modelName. Learning parts are highlighted in yellow, and frozen parts are shown in gray.
    Step 1. Initial tuning phase where the LM is fine-tuned to generate task-specific embeddings. This step is optional.
    Step 2. Training \modelName to transform \( e(d_e) \) into \( e(\mathcal{T}(d_e)) \). \( d_e \) and \( \mathcal{T}(d_e) \) carry the same meaning. 
    Step 3. Fine-tuning the task using the trained \modelName. The model is trained with English data.
    }
    \label{fig:steps}

\end{figure*}

\subsection{Fine-tuning with \modelName} \label{sec:LAtuning}
Figure~\ref{fig:steps} illustrates the procedure of task fine-tuning using \modelName. 
The training process can be divided into three main steps.

\paragraph{Step 1: Tuning for Embedding}
In this step, the LM is fine-tuned with task-specific data to generate task-suitable embeddings. Although optional, omitting this step led to performance degradation.

\paragraph{Step 2: \modelName Training}
In this step, \modelName is trained to take \( e(d_e) \) as input and produce \( e(\mathcal{T}(d_e)) \) as output. To maintain consistent embeddings, \modelName is also trained to produce \( e(\mathcal{T}(d_e)) \) when  \( \mathcal{T}(d_e) \) is given as input. The LM used in this step is the one fine-tuned in Step 1 and is frozen to prevent any changes in the embeddings. Mean Squared Error (MSE) is used as the loss function.

\paragraph{Step 3: Fine-tuning for Task}
In this step, the model with \modelName is fine-tuned for the target task. 
\modelName is placed between the LM and the task head, adjusting the English embeddings to match those of the target language before passing them to the task head. 
The task head is newly defined in the same form as the LM's task head.

\paragraph{Training Data for \modelName}\label{sec:sepData}
Training \modelName requires parallel corpora to align embeddings between languages. We used a subset of the English data intended for transfer learning in our benchmark tasks, translating it using the GPT-4 Turbo API, known for its high performance in machine translation (MT). We evaluated \modelName’s performance using datasets of 5,000 to 20,000 samples, increasing in increments of 5,000. For fairness, the data used to train \modelName was kept separate from the fine-tuning data, ensuring it handles the fine-tuning data as unseen.

\subsection{Experiment Design}
We designed a two-stage experiment using benchmark tests to evaluate \modelName's effectiveness.

\paragraph{Benchmark Evaluation}
To assess \modelName’s value as a data processing method, we performed the following experiments:  
(1) We generated benchmark training data using various data construction and processing methods, including \modelName.  
(2) We compared the resources required for data construction and the resulting model performance.  
We used a test set of 2,000 samples in the target language for evaluation.

\paragraph{Ablation Study}
Two ablation studies were conducted to verify that the performance improvements were due to \modelName.  
(1) We removed the \modelName layer and measured performance to confirm its impact.  
(2) We used identical data for \modelName training and task tuning to ensure that performance differences were not due to data variation. This also demonstrates that, in practice, \modelName does not require splitting valuable translated data.

\subsection{Experiment Setting}
\subsubsection{Target Languages}
Korean was chosen as the primary target language due to our familiarity. To assess the model's generalization capability, we included Japanese and Chinese, which are linguistically distinct from English, the source language for transfer learning. Detailed characteristics are provided in Appendix~\ref{Ap:Language}.

\begin{table}[]
\begin{adjustbox}{width=\linewidth}

\begin{tabular}{c|c|cccc}
\hline
\multirow{2}{*}{} & \multirow{2}{*}{criterion} & \multicolumn{4}{c}{Language}             \\ \cline{3-6} 
                  &                            & English  & Korean   & Japanese & Chinese \\ \hline
SA                & F1 score                   & \makecell[c]{GLUE-SA\\~\citep{wang2018glue}}  & \makecell[c]{NSMC\\~\citep{Park:2016}}     & -        & \makecell[c]{Douban~\footnote{https://github.com/kakayuw/Sentiment-Analysis-Based-on-Douban-Movie-Critics}}  \\
NLI               & Accuracy                   & \makecell[c]{GLUE-NLI\\~\citep{wang2018glue}} & \makecell[c]{KLUE-NLI\\\citep{park2021klue}} & \makecell[c]{JNLI\\~\citep{kurihara-etal-2022-jglue}}     & \makecell[c]{ocNLI\\~\citep{ocnli}}   \\ \hline
\end{tabular}
\end{adjustbox}
\caption{Information of the benchmark datasets.}\label{tab:benchInfo}
\end{table}
\subsubsection{Evaluation Tasks}
The experiments utilized benchmark tasks available across the three target languages and English. Further details on each task can be found in Table~\ref{tab:benchInfo} and Appendix~\ref{ap:Benchmarks}.

\paragraph{Sentiment Analysis (SA)}
This task involves classifying a sentence based on its sentiment as either positive or negative.

\paragraph{Natural Language Inference (NLI)}
This task involves determining the relationship between a premise and a hypothesis, categorizing it as entailment, contradiction, or neutral.

\subsubsection{Baselines}
The goal of \modelName is to enhance transfer learning performance to match models trained on target language data, reducing data collection costs and efforts.
We set various types of data as baselines for our experiments. 
These models follow the typical LM training procedure rather than \modelName's training procedure. 
Detailed information on each baseline and their generation methods is provided in Appendix~\ref{ap:baselines}.

\paragraph{Native Data (Korean, Japanese, Chinese)}
This dataset, curated by native speakers, achieves the highest performance but at a significant collection cost. It serves as the ideal reference for evaluating data construction and transfer learning approaches.

\paragraph{English Data (English)} \label{ph:en}
As the fundamental source for transfer learning, English data is cost-effective and readily available, albeit with lower performance compared to native data. This dataset is also used to create the translated datasets described below, ensuring consistent semantic content across all datasets.

\paragraph{High-Quality NMT (NMT\texorpdfstring{\textsubscript{GPT-4}}{NMTGPT-4})}
Well-translated data, by humans or LLMs, is effective for data acquisition. 
We translated the EN data using GPT-4 Turbo, known for high performance in this task~\citep{jiao2023chatgpt}. Since \modelName requires translation data for training, TR-GPT became the practical performance goal.

\paragraph{Local-Tuned NMT (NMT\texorpdfstring{\textsubscript{mT5}}{NMTmT5}})
To assess the effectiveness of \modelName compared to leveraging machine translation, we employ Google’s mT5-base~\citep{xue2021mt5} to translate the English data. The mT5 translator was fine-tuned using the parallel corpus employed for training \modelName, with the size of corpus indicated accordingly.

\begin{table}[]
\begin{adjustbox}{width=\columnwidth}
\begin{tabular}{c|cccc}
\hline
                & Opt. & Loss func. & l-rate&batch size       \\ \hline
\modelName       & adamW     & MSE        & \(1 \times 10^{-5}\)       &16 \\
benchmark task  & adamW     & X-entropy  & \(1 \times 10^{-4}\)       &16\\
mT5 translation & adamW     & MSE        & \(1 \times 10^{-5}\)      &8 \\\hline
\end{tabular}
\end{adjustbox}
\caption{Information on Hyperparameters}
\label{tb:hyperParam}
\end{table}

\begin{table*}[]
\begin{adjustbox}{width=\linewidth}
\begin{tabular}{cc|ccccccc|ccccccc}
\hline
                                                  &                & \multicolumn{7}{c|}{SA (F1 Score)}                    & \multicolumn{7}{c}{NLI (Accuracy)}                    \\ \hline
Methods                  & Tr. Data                &  5K    & 10K   & 15K   & 20K   & 30K   & 40K   & 50K   & 5K    & 10K   & 15K   & 20K   & 30K   & 40K   & 50K   \\ \hline
\multicolumn{2}{l|}{baselines}                                       &       &       &       &       &       &       &       &       &       &       &       &       &       &       \\ \cline{1-2}
English                        & English                               & 0.742 & 0.742 & 0.747 & 0.748 & 0.749 & 0.753 & 0.754 & 0.688 & 0.709 & 0.713 & 0.724 & 0.723 & 0.724 & 0.731 \\
Korean                        & Korean                               & 0.830  & 0.851 & 0.849 & 0.857 & -     & -     & -     & 0.739 & 0.792 & 0.795 & 0.812 & -     & -     & -     \\
NMT\textsubscript{GPT-4}                   & Korean                               & 0.772 & 0.788 & 0.782 & \cellcolor[HTML]{8D999E}0.787 & -     & -     & -     & 0.726 & 0.785 & 0.766 & \cellcolor[HTML]{8D999E}0.774 & -     & -     & -     \\
NMT\textsubscript{mT5-20K}                & Korean                                & 0.680  & 0.657 & 0.681 & 0.634 & -     & -     & -     & 0.614 & 0.598 & 0.650  & 0.579 & -     & -     & -     \\ \hline
\multicolumn{2}{l|}{LangAlign}                                    &       &       &       &       &       &       &       &       &       &       &       &       &       &       \\ \cline{1-2}
AE 5K                      & English                             & 0.764 & 0.780  & \textbf{0.790}  & \textbf{0.797} & \underline{0.801} & \underline{0.802} & \underline{0.806} & 0.714 & 0.764 & 0.718 & \textbf{0.783} & 0.766 & \underline{0.788} & \underline{0.777} \\
AE 15K                     & English                            & \textbf{0.781} & \textbf{0.790}  & \textbf{0.795} & \textbf{0.803} & \underline{0.811} & \underline{0.811} & \underline{0.820}  & 0.700   & 0.775 & 0.744 & 0.756 & \underline{0.799} & \underline{0.799} & \underline{0.801} \\
FC 5K                      & English                             & 0.744 & 0.763 & 0.775 & 0.779 & 0.777 & \underline{0.789} & \underline{0.784} & 0.666 & 0.737 & 0.719 & 0.736 & 0.764 & 0.754 & \underline{0.776} \\
FC 15K                     & English                             & 0.766 & 0.768 & \textbf{0.782} & \textbf{0.795} & \underline{0.801} & \underline{0.813} & \underline{0.810}  & 0.680  & 0.710  & \textbf{0.776} & 0.762 & \underline{0.781} & \underline{0.778} & \underline{0.781} \\ \hline

\end{tabular}
\end{adjustbox}
\caption{Performance of models using \modelName. ``Methods'' shows how training data was generated, and ``Tr.Data'' indicates the task tuning language. Subscripts in ``Methods'' (e.g., NMT\textsubscript{mT5-20K}, AE\textsubscript{5K}) denote the amount of GPT-4 translated data used. \textbf{Bold}: \modelName trained on English outperforms NMT\textsubscript{GPT-4} trained on the same amount of Korean data. \underline{Underline}: \modelName catches up by leveraging abundant, low-cost English data.
}\label{tab:mainResult}
\end{table*}
\subsubsection{Implementation Details}
Our experiments were conducted using Python, PyTorch, and the Transformers library in an A100 40GB environment. Table~\ref{tb:hyperParam} summarizes the parameters used. We reserved 10\% of the training data for validation and applied early stopping with a patience of 10 epochs. OS and library details are provided in Appendix~\ref{ap:impDetail}.

\paragraph{XLM-RoBERTa-base}
We used XLM-RoBERTa~\citep{DBLP:journals/corr/abs-1911-02116} as the LM, considering its robustness across multilingual benchmarks. Developed by Facebook AI, this model is pre-trained on data from over 100 languages, supporting effective transfer learning. The base model has 125M parameters, 12 layers, a sequence length of 512 tokens, and outputs 768-dimensional embeddings.

\section{Experimental Result \& Discussion}
Table~\ref{tab:mainResult} presents the experimental results. The "Tr.Data" column indicates the language used for task-specific tuning, and subscripts in ``Methods'' (e.g., NMT\textsubscript{mT5}, AE 5K) denote the amount of GPT-4 translated data used.

We primarily used datasets up to 20K samples for evaluation, with English data and \modelName extended to 50K samples considering the abundance of English resources.
Results are shown in increments of 5K for better readability, although experiments were conducted at intervals of 1K.

The parallel corpus sizes for \modelName and NMT\textsubscript{mT5} training were set at 5K, 10K, 15K, and 20K. Only a subset is reported for readability, with complete results provided in the Appendix~\ref{ap:resultFull}.

\paragraph{vs. English Data} The \modelName-applied model consistently outperformed the model trained on English data alone, demonstrating the effectiveness of \modelName in enhancing tuning performance.

\paragraph{vs. NMT\textsubscript{mT5}} The data translated by the mT5 model, fine-tuned with less than 20K parallel corpus, showed lower performance than the model trained directly on English data. This suggests that when only a small amount of parallel data (e.g., 20K samples) is available and computational resources are limited, \modelName can be a more effective approach than direct NMT.

\paragraph{vs. Korean and NMT\textsubscript{GPT-4}} 
Models trained on Korean data alone generally performed better for datasets of the same size. However, by leveraging the abundance of English data, \modelName was able to surpass the performance of the Korean data models. For instance, in the NLI task, the AE-5K model, which used 5K translated samples, scored 0.788 when combined with 40K English samples—significantly outperforming the 0.739 score of the model trained on 5K Korean data.

For comparison with NMT\textsubscript{GPT-4}, we used the following indicators:
(1) \textbf{Bold font} indicates cases where the \modelName model outperformed models trained on an equivalent amount of translated data.
(2) \underline{Underlined font} highlights cases where the \modelName model leveraged the abundance of English data to surpass the performance of the NMT\textsubscript{GPT-4} model trained with the largest amount of translated data (20K case, highlighted in gray). For example, in the SA task, the AE-5K model fine-tuned with 30K English samples (0.811) outperformed the model trained with 20K translated samples—even though it used only 5K translated data, which is significantly more expensive to obtain.

These results indicate that \modelName can effectively enhance model performance by utilizing the abundance and cost-efficiency of English data, even when high-cost translation data is limited. This highlights the cost-efficiency of \modelName in achieving competitive performance. Additionally, its ability to closely approximate the performance of models trained on native data showcases its industrial value in estimating the lower bounds of direct data construction.

\begin{figure*}[h]
    \centering
    \includegraphics[width=\linewidth]{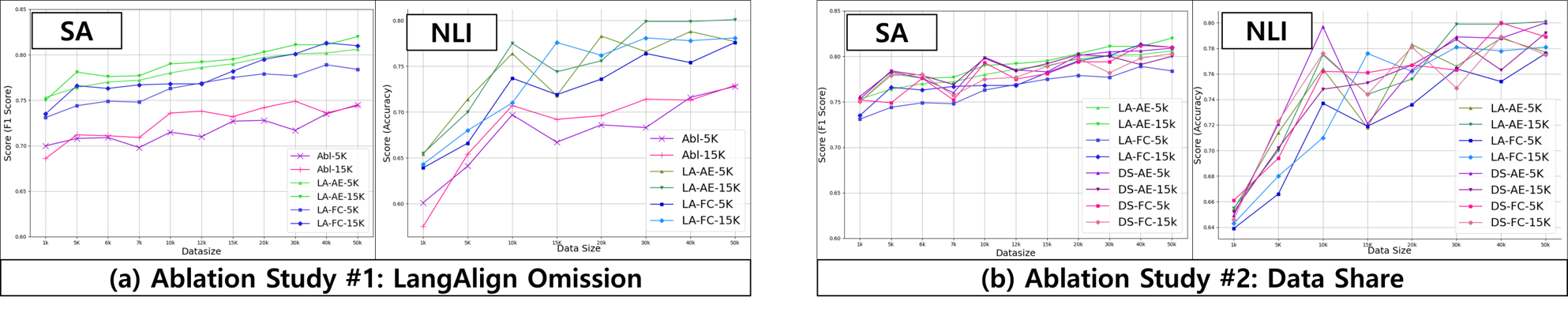}
    \caption{Ablation study results. (a) Performance of models with the \modelName layer removed, (b) Performance of models using the same data for both \modelName training and task-specific tuning.}
    \label{fig:ablResult}    
\end{figure*}

\begin{figure}[h]
    \centering
    \includegraphics[width=0.7\linewidth]{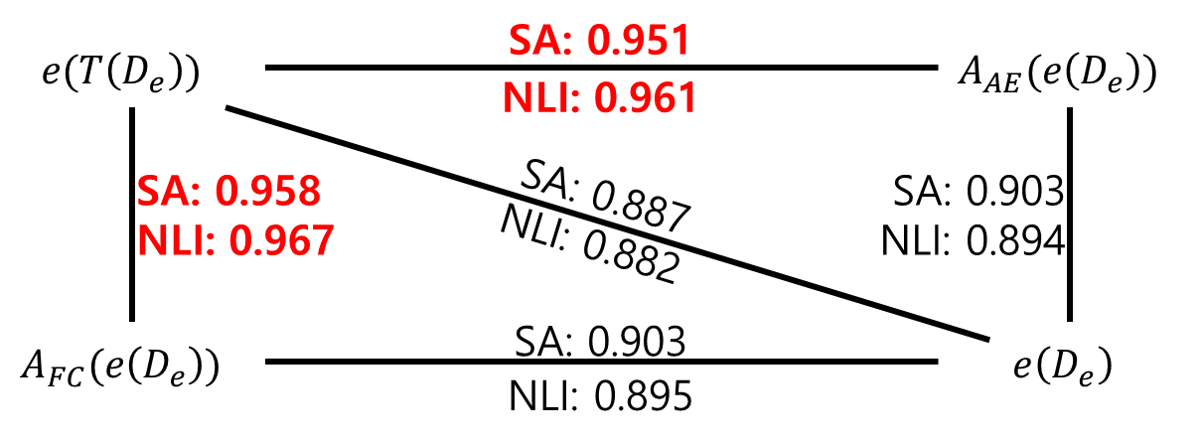} 
    \caption{Verification of \modelName’s embedding transformation capability using cosine similarity}

    \label{fig:cosine}
\end{figure}
\begin{figure}[h]
    \centering

    \includegraphics[width=0.6\linewidth]{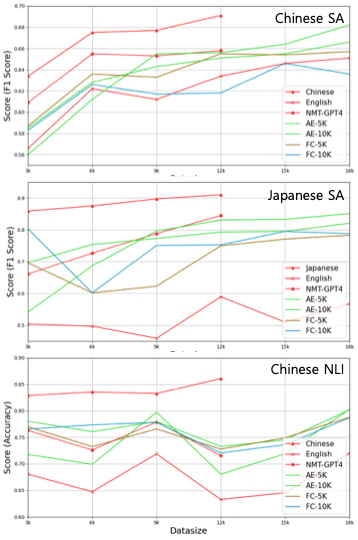}
    \caption{Language generalizability evaluation results}
    \label{fig:foreign}

\end{figure}

\subsection{Ablation Study} 
We conducted two ablation studies to validate the effectiveness of \modelName. Instead of focusing on specific numerical values, we analyzed trends and patterns, which are presented in Figure~\ref{fig:ablResult} as a plot chart. Detailed results can be found in Appendix~\ref{ap:resultFull}.

\paragraph{\modelName Omission}
Figure~\ref{fig:ablResult}(a) shows that models following the \modelName training procedure but without the \modelName layer (\textcolor[HTML]{9400D3}{Abl-5K} \& \textcolor[HTML]{FF1493}{Abl-15K}) exhibit significantly lower performance compared to models with the \modelName layer applied (\textcolor[HTML]{6B8E23}{AE-5K}, \textcolor[HTML]{2E8B57}{AE-15K}, \textcolor[HTML]{0000CD}{FC-15K}, and \textcolor[HTML]{1E90FF}{FC-15K}).

\paragraph{Data Share}
Figure~\ref{fig:ablResult}(b) illustrates the results when the same data is used for both \modelName training and task-specific tuning. Models using shared data show performance nearly identical to those using independent datasets, which contrasts with the performance drop observed in the \modelName omission case.

These two ablation studies confirm that the performance improvements observed in models applying \modelName are solely attributed to the \modelName framework.

\subsection{Additional Experiments}
\subsubsection{Cosine Similarity}
To verify \modelName's effectiveness in transforming \( e(d_e) \) to \( e(\mathcal{T}(d_e)) \), we measured cosine similarity using 20k EN and TR-GPT data. 
Figure~\ref{fig:cosine} shows the cosine similarity results when \modelName is trained with 20k parallel corpora.
The embeddings processed by \modelName are closer to translated data embeddings than to English data embeddings (SA: 0.958 vs 0.903, NLI: 0.967 vs 0.894). 
Detailed cosine similarity results for various parallel corpus sizes are in Appendix~\ref{ap:cosineDetail}.

\subsubsection{Experiment with Other Languages}
We evaluated \modelName on Japanese and Chinese to assess its linguistic generalization capabilities. Figure~\ref{fig:foreign} presents the results for Japanese NLI and Chinese SA and NLI tasks. \modelName performed comparably to NMT\textsubscript{GPT-4} and surpassed it as data size increased.

\begin{table}[]
\centering
\begin{adjustbox}{width=0.9\columnwidth}
\begin{tabular}{l|cc|ccc}
\hline
           & \multicolumn{2}{c|}{SA} & \multicolumn{3}{c}{NLI} \\ \cline{2-6}
Scale $\rightarrow$           & 20K         & 50K       & 20K    & 50K    & 300K  \\ \hline
English         & 0.748       & 0.754     & 0.724  & 0.731  & 0.761 \\
TR-GPT     & 0.787       & -         & 0.774  & -      & -     \\ \cdashline{1-6}\noalign{\vskip 0.5ex}
Rev-AE 20K & 0.766       & 0.797     & 0.762 & 0.766 & 0.812 \\
Rev-FC 20K & 0.762      & 0.777     & 0.749 & 0.752  & 0.819 \\ \hline
\end{tabular}
\end{adjustbox}
\caption{
Performance of the English Model with \revModelNM and the Korean-trained Model}

\label{tab:revSimple}
\end{table}

\subsubsection{Reversed \modelName}
We explored the reverse approach of \modelName, utilizing English data as the training set and transforming Korean inputs into the English domain. Specifically, we reversed Step 2 from Section~\ref{sec:LAtuning} and applied it to a model trained on English data.

The results are shown in Table~\ref{tab:revSimple}. \revModelNM outperformed standard transfer learning (English) with the same amount of data and showed slightly lower performance than NMT\textsubscript{GPT-4}. However, at 50K, \revModelNM achieved performance comparable to NMT\textsubscript{GPT-4}. Notably, when scaling up to 300K, \revModelNM surpassed NMT\textsubscript{GPT-4} in some cases, benefiting significantly from the larger English dataset.

This experiment demonstrates that \modelName can effectively leverage English data as training data for Korean modeling and also shows that transforming Korean inputs to be compatible with an English model (\revModelNM) is a viable approach. The full experimental results of \revModelNM are provided in Appendix~\ref{ap:revFull}.

\begin{table}[]
\begin{adjustbox}{width=\linewidth}
\begin{tabular}{c|cccc|cccc}
\hline
              & \multicolumn{4}{c|}{SA}       & \multicolumn{4}{c}{NLI}       \\ \hline
datasize      & 5K    & 10K   & 15K   & 20K   & 5K    & 10K   & 15K   & 20K   \\ \hline
LoRA-TA       & 0.746 & 0.759 & 0.759 & 0.753 & 0.714 & 0.733 & 0.741 & 0.738 \\
Unsup. SimSCE & 0.614 & 0.541 & 0.465 & 0.597 & 0.688 & 0.701 & 0.699 & 0.700 \\
\hline
\end{tabular}
\end{adjustbox}
\caption{Performance of LoRA-TA and SimSCE models}\label{tab:loraBrief}
\end{table}
\subsection{Comparison with Other Architectures}  
Our primary focus has been on the cost-efficiency of data construction using \modelName. In addition, we briefly evaluated performance comparisons with other architectures that share some conceptual similarities, such as LoRA Task Adaptation (LoRA-TA,~\citealp{hu2021lora}) and Unsupervised-SimSCE(SimSCE,~\citealp{gao2021simcse}). The results are summarized in Table~\ref{tab:loraBrief}. In short, the two models did not show significant benefits under our experimental conditions. This can be attributed to using a LM capable of tuning all parameters and the lack of sufficient data for enhancing embeddings. 
To maintain consistency, we have briefly summarized this section; detailed procedures and analyses are provided in Appendix~\ref{ap:otherArchitectures}.

\section{Conclusion}
We introduced \modelName to enhance the efficiency of transfer learning.
\modelName transforms English data embeddings into target language embeddings, offering a cost-effective alternative to using translated data, while leveraging the the richness of English data. 
\modelName demonstrated language generalizability, and \revModelNM showed that target language input can be effectively converted into the embedding space of English data, suggesting potential for transfer inference. 
\modelName holds industrial value by reducing the time and cost of data construction while enhancing performance.

\newpage

\bibliography{custom}

\begin{thebibliography}{30}
\providecommand{\natexlab}[1]{#1}

\bibitem[{Achiam et~al.(2023)Achiam, Adler, Agarwal, Ahmad, Akkaya, Aleman, Almeida, Altenschmidt, Altman, Anadkat et~al.}]{achiam2023gpt}
Josh Achiam, Steven Adler, Sandhini Agarwal, Lama Ahmad, Ilge Akkaya, Florencia~Leoni Aleman, Diogo Almeida, Janko Altenschmidt, Sam Altman, Shyamal Anadkat, et~al. 2023.
\newblock Gpt-4 technical report.
\newblock \emph{arXiv preprint arXiv:2303.08774}.

\bibitem[{Bubeck et~al.(2023)Bubeck, Chandrasekaran, Eldan, Gehrke, Horvitz, Kamar, Lee, Lee, Li, Lundberg et~al.}]{bubeck2023sparks}
S{\'e}bastien Bubeck, Varun Chandrasekaran, Ronen Eldan, Johannes Gehrke, Eric Horvitz, Ece Kamar, Peter Lee, Yin~Tat Lee, Yuanzhi Li, Scott Lundberg, et~al. 2023.
\newblock Sparks of artificial general intelligence: Early experiments with gpt-4.
\newblock \emph{arXiv preprint arXiv:2303.12712}.

\bibitem[{Conneau et~al.(2019{\natexlab{a}})Conneau, Khandelwal, Goyal, Chaudhary, Wenzek, Guzm{\'a}n, Grave, Ott, Zettlemoyer, and Stoyanov}]{conneau2019unsupervised}
Alexis Conneau, Kartikay Khandelwal, Naman Goyal, Vishrav Chaudhary, Guillaume Wenzek, Francisco Guzm{\'a}n, Edouard Grave, Myle Ott, Luke Zettlemoyer, and Veselin Stoyanov. 2019{\natexlab{a}}.
\newblock Unsupervised cross-lingual representation learning at scale.
\newblock \emph{arXiv preprint arXiv:1911.02116}.

\bibitem[{Conneau et~al.(2019{\natexlab{b}})Conneau, Khandelwal, Goyal, Chaudhary, Wenzek, Guzm{\'{a}}n, Grave, Ott, Zettlemoyer, and Stoyanov}]{DBLP:journals/corr/abs-1911-02116}
Alexis Conneau, Kartikay Khandelwal, Naman Goyal, Vishrav Chaudhary, Guillaume Wenzek, Francisco Guzm{\'{a}}n, Edouard Grave, Myle Ott, Luke Zettlemoyer, and Veselin Stoyanov. 2019{\natexlab{b}}.
\newblock \href {https://arxiv.org/abs/1911.02116} {Unsupervised cross-lingual representation learning at scale}.
\newblock \emph{CoRR}, abs/1911.02116.

\bibitem[{Devlin et~al.(2018)Devlin, Chang, Lee, and Toutanova}]{devlin2018bert}
Jacob Devlin, Ming-Wei Chang, Kenton Lee, and Kristina Toutanova. 2018.
\newblock Bert: Pre-training of deep bidirectional transformers for language understanding.
\newblock \emph{arXiv preprint arXiv:1810.04805}.

\bibitem[{Gaim et~al.(2023)Gaim, Yang, Park, and Park}]{gaim2023question}
Fitsum Gaim, Wonsuk Yang, Hancheol Park, and Jong~C Park. 2023.
\newblock Question-answering in a low-resourced language: Benchmark dataset and models for tigrinya.
\newblock In \emph{Proceedings of the 61st Annual Meeting of the Association for Computational Linguistics (Volume 1: Long Papers)}, pages 11857--11870.

\bibitem[{Gao et~al.(2023)Gao, Zhang, He, Wu, and Wang}]{gao2023learning}
Pengzhi Gao, Liwen Zhang, Zhongjun He, Hua Wu, and Haifeng Wang. 2023.
\newblock Learning multilingual sentence representations with cross-lingual consistency regularization.
\newblock \emph{arXiv preprint arXiv:2306.06919}.

\bibitem[{Gao et~al.(2021)Gao, Yao, and Chen}]{gao2021simcse}
Tianyu Gao, Xingcheng Yao, and Danqi Chen. 2021.
\newblock Simcse: Simple contrastive learning of sentence embeddings.
\newblock \emph{arXiv preprint arXiv:2104.08821}.

\bibitem[{Houlsby et~al.(2019)Houlsby, Giurgiu, Jastrzebski, Morrone, De~Laroussilhe, Gesmundo, Attariyan, and Gelly}]{houlsby2019parameter}
Neil Houlsby, Andrei Giurgiu, Stanislaw Jastrzebski, Bruna Morrone, Quentin De~Laroussilhe, Andrea Gesmundo, Mona Attariyan, and Sylvain Gelly. 2019.
\newblock Parameter-efficient transfer learning for nlp.
\newblock In \emph{International conference on machine learning}, pages 2790--2799. PMLR.

\bibitem[{Hu et~al.(2021)Hu, Shen, Wallis, Allen-Zhu, Li, Wang, Wang, and Chen}]{hu2021lora}
Edward~J. Hu, Yelong Shen, Phillip Wallis, Zeyuan Allen-Zhu, Yuanzhi Li, Shean Wang, Lu~Wang, and Weizhu Chen. 2021.
\newblock \href {https://openreview.net/forum?id=nZeVKeeFYf9} {Lora: Low-rank adaptation of large language models}.
\newblock \emph{OpenReview}.
\newblock ICLR 2022.

\bibitem[{Hu et~al.(2020)Hu, Richardson, Xu, Li, Kuebler, and Moss}]{ocnli}
Hai Hu, Kyle Richardson, Liang Xu, Lu~Li, Sandra Kuebler, and Larry Moss. 2020.
\newblock \href {https://arxiv.org/abs/2010.05444} {Ocnli: Original chinese natural language inference}.
\newblock In \emph{Findings of EMNLP}.

\bibitem[{Jiang et~al.(2024)Jiang, Drummond, and Cohn}]{jiang2024pre}
Fan Jiang, Tom Drummond, and Trevor Cohn. 2024.
\newblock Pre-training cross-lingual open domain question answering with large-scale synthetic supervision.
\newblock \emph{arXiv preprint arXiv:2402.16508}.

\bibitem[{Jiao et~al.(2023)Jiao, Wang, Huang, Wang, Shi, and Tu}]{jiao2023chatgpt}
Wenxiang Jiao, Wenxuan Wang, Jen-tse Huang, Xing Wang, Shuming Shi, and Zhaopeng Tu. 2023.
\newblock Is chatgpt a good translator? yes with gpt-4 as the engine.
\newblock \emph{arXiv preprint arXiv:2301.08745}.

\bibitem[{Joshi et~al.(2020)Joshi, Santy, Budhiraja, Bali, and Choudhury}]{joshi2020state}
Pratik Joshi, Sebastin Santy, Amar Budhiraja, Kalika Bali, and Monojit Choudhury. 2020.
\newblock The state and fate of linguistic diversity and inclusion in the nlp world.
\newblock \emph{arXiv preprint arXiv:2004.09095}.

\bibitem[{Kurihara et~al.(2022)Kurihara, Kawahara, and Shibata}]{kurihara-etal-2022-jglue}
Kentaro Kurihara, Daisuke Kawahara, and Tomohide Shibata. 2022.
\newblock \href {https://aclanthology.org/2022.lrec-1.317} {{JGLUE}: {J}apanese general language understanding evaluation}.
\newblock In \emph{Proceedings of the Thirteenth Language Resources and Evaluation Conference}, pages 2957--2966, Marseille, France. European Language Resources Association.

\bibitem[{Moosavi et~al.(2022)Moosavi, Delfosse, Kersting, and Gurevych}]{moosavi-etal-2022-adaptable}
Nafise Moosavi, Quentin Delfosse, Kristian Kersting, and Iryna Gurevych. 2022.
\newblock \href {https://doi.org/10.18653/v1/2022.naacl-main.274} {Adaptable adapters}.
\newblock In \emph{Proceedings of the 2022 Conference of the North American Chapter of the Association for Computational Linguistics: Human Language Technologies}, pages 3742--3753, Seattle, United States. Association for Computational Linguistics.

\bibitem[{Norman(1988)}]{norman1988chinese}
Jerry Norman. 1988.
\newblock \emph{Chinese}.
\newblock Cambridge University Press.

\bibitem[{Park(2016)}]{Park:2016}
Lucy Park. 2016.
\newblock Naver sentiment movie corpus.

\bibitem[{Park et~al.(2021)Park, Moon, Kim, Cho, Han, Park, Song, Kim, Song, Oh et~al.}]{park2021klue}
Sungjoon Park, Jihyung Moon, Sungdong Kim, Won~Ik Cho, Jiyoon Han, Jangwon Park, Chisung Song, Junseong Kim, Yongsook Song, Taehwan Oh, et~al. 2021.
\newblock Klue: Korean language understanding evaluation.
\newblock \emph{arXiv preprint arXiv:2105.09680}.

\bibitem[{Ramsey(1987)}]{ramsey1987languages}
S~Robert Ramsey. 1987.
\newblock \emph{The languages of China}.
\newblock Princeton University Press Princeton, NJ.

\bibitem[{Schuster et~al.(2018)Schuster, Gupta, Shah, and Lewis}]{schuster2018cross}
Sebastian Schuster, Sonal Gupta, Rushin Shah, and Mike Lewis. 2018.
\newblock Cross-lingual transfer learning for multilingual task oriented dialog.
\newblock \emph{arXiv preprint arXiv:1810.13327}.

\bibitem[{Shibatani(1990)}]{shibatani1990languages}
Masayoshi Shibatani. 1990.
\newblock \emph{The languages of Japan}.
\newblock Cambridge University Press.

\bibitem[{Sohn(2005)}]{sohn2005korean}
Ho-min Sohn. 2005.
\newblock \emph{Korean language in culture and society}.
\newblock University of Hawaii press.

\bibitem[{Tan et~al.(2022)Tan, Heffernan, Schwenk, and Koehn}]{tan2022multilingual}
Weiting Tan, Kevin Heffernan, Holger Schwenk, and Philipp Koehn. 2022.
\newblock Multilingual representation distillation with contrastive learning.
\newblock \emph{arXiv preprint arXiv:2210.05033}.

\bibitem[{Tsujimura(2013)}]{tsujimura2013introduction}
Natsuko Tsujimura. 2013.
\newblock \emph{An introduction to Japanese linguistics}.
\newblock John Wiley \& Sons.

\bibitem[{Wang et~al.(2018)Wang, Singh, Michael, Hill, Levy, and Bowman}]{wang2018glue}
Alex Wang, Amanpreet Singh, Julian Michael, Felix Hill, Omer Levy, and Samuel~R Bowman. 2018.
\newblock Glue: A multi-task benchmark and analysis platform for natural language understanding.
\newblock \emph{arXiv preprint arXiv:1804.07461}.

\bibitem[{Wang et~al.(2022)Wang, Mishra, Alipoormolabashi, Kordi, Mirzaei, Arunkumar, Ashok, Dhanasekaran, Naik, Stap et~al.}]{wang2022super}
Yizhong Wang, Swaroop Mishra, Pegah Alipoormolabashi, Yeganeh Kordi, Amirreza Mirzaei, Anjana Arunkumar, Arjun Ashok, Arut~Selvan Dhanasekaran, Atharva Naik, David Stap, et~al. 2022.
\newblock Super-naturalinstructions: Generalization via declarative instructions on 1600+ nlp tasks.
\newblock \emph{arXiv preprint arXiv:2204.07705}.

\bibitem[{Xue et~al.(2020)Xue, Constant, Roberts, Kale, Al-Rfou, Siddhant, Barua, and Raffel}]{xue2020mt5}
Linting Xue, Noah Constant, Adam Roberts, Mihir Kale, Rami Al-Rfou, Aditya Siddhant, Aditya Barua, and Colin Raffel. 2020.
\newblock mt5: A massively multilingual pre-trained text-to-text transformer.
\newblock \emph{arXiv preprint arXiv:2010.11934}.

\bibitem[{Xue et~al.(2021)Xue, Constant, Roberts, Kale, Al-Rfou, Siddhant, Barua, and Raffel}]{xue2021mt5}
Linting Xue, Noah Constant, Adam Roberts, Mihir Kale, Rami Al-Rfou, Aditya Siddhant, Aditya Barua, and Colin Raffel. 2021.
\newblock \href {https://doi.org/10.18653/v1/2021.naacl-main.41} {mt5: A massively multilingual pre-trained text-to-text transformer}.
\newblock In \emph{Proceedings of the 2021 Conference of the North American Chapter of the Association for Computational Linguistics: Human Language Technologies}, pages 483--498. Association for Computational Linguistics.

\bibitem[{Yeon and Brown(2013)}]{yeon2013korean}
Jaehoon Yeon and Lucien Brown. 2013.
\newblock \emph{Korean: A comprehensive grammar}.
\newblock Routledge.

\end{thebibliography}

\newpage
\appendix
\section{\modelName Details} \label{ap:modelDetail}
Listing ~\ref{code:simplefc} and ~\ref{code:AE} and Figure  show structure of \modelName.

\begin{figure*}[t]
\begin{lstlisting}[language=Python, caption=FC \modelName, label=code:simplefc, frame=single, breaklines=true]
class SimpleFC(nn.Module):
    def __init__(self, input_dim, output_dim):
        super(SimpleFC, self).__init__()
        self.fc1 = nn.Linear(input_dim, 1024)
        self.fc2 = nn.Linear(1024, 512)
        self.fc3 = nn.Linear(512, 256)
        self.fc4 = nn.Linear(256, 512)
        self.fc5 = nn.Linear(512, 1024)
        self.fc6 = nn.Linear(1024, output_dim)
        self.dropout = nn.Dropout(0.5)

    def forward(self, x):
        x = torch.relu(self.fc1(x))
        x = self.dropout(x)
        x = torch.relu(self.fc2(x))
        x = self.dropout(x)
        x = torch.relu(self.fc3(x))
        x = torch.relu(self.fc4(x))
        x = self.dropout(x)
        x = torch.relu(self.fc5(x))
        x = self.fc6(x)
        return x

input_dim = 768
output_dim = 768

model = SimpleFC(input_dim, output_dim)

criterion = nn.MSELoss()
optimizer = optim.Adam(model.parameters(), lr=0.0001)
\end{lstlisting}
\end{figure*}\label{listing:FC}

\begin{figure*}[t]
\begin{lstlisting}[language=Python, caption=AE \modelName, label=code:AE, frame=single, breaklines=true]
class Encoder(nn.Module):
        def __init__(self, input_dim, hidden_dim):
            super(Encoder, self).__init__()
            self.fc1 = nn.Linear(input_dim, hidden_dim)
            self.fc2 = nn.Linear(hidden_dim, hidden_dim // 2)
            self.fc3 = nn.Linear(hidden_dim // 2, hidden_dim // 4)

        def forward(self, x):
            x = torch.relu(self.fc1(x))
            x = torch.relu(self.fc2(x))
            x = torch.relu(self.fc3(x))
            return x

    class Decoder(nn.Module):
        def __init__(self, hidden_dim, output_dim):
            super(Decoder, self).__init__()
            self.fc1 = nn.Linear(hidden_dim // 4, hidden_dim // 2)
            self.fc2 = nn.Linear(hidden_dim // 2, hidden_dim)
            self.fc3 = nn.Linear(hidden_dim, output_dim)

        def forward(self, x):
            x = torch.relu(self.fc1(x))
            x = torch.relu(self.fc2(x))
            x = self.fc3(x)
            return x

    class AutoEncoder(nn.Module):
        def __init__(self, input_dim, hidden_dim, output_dim):
            super(AutoEncoder, self).__init__()
            self.encoder = Encoder(input_dim, hidden_dim)
            self.decoder = Decoder(hidden_dim, output_dim)

        def forward(self, x):
            x = self.encoder(x)
            x = self.decoder(x)
            return x

    input_dim = 768
    hidden_dim = 512
    output_dim = 768

    model = AutoEncoder(input_dim, hidden_dim, output_dim)

    criterion = nn.MSELoss()
    optimizer = optim.Adam(model.parameters(), lr=0.0001)
\end{lstlisting}
\end{figure*}\label{listing:AE}

\begin{figure*}
    \centering[h]
    \includegraphics[width=\linewidth]{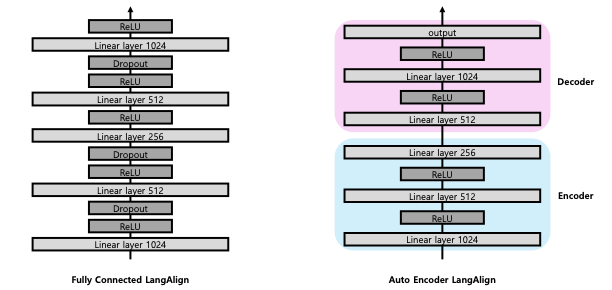}
    \caption{Layer structure of \modelName}
\end{figure*}

\section{Korean, Chinese, and Japanese: Linguistic Features, Word Order, and Writing Systems}\label{Ap:Language}
\subsection{Korean}
\paragraph{Word Order}: Korean primarily uses Subject-Object-Verb (SOV) word order. For example, "나(I)는 사과(apple)를 먹는다(eat)" (I eat an apple).
\paragraph{Grammar}: Korean employs particles to indicate grammatical relationships between words. It also has a well-developed system of honorifics.
Writing System: Hangul is a phonetic alphabet where each character represents a sound.
\paragraph{Differences from English}: English uses Subject-Verb-Object (SVO) word order and prepositions, while Korean uses particles and a different word order~\citep{sohn2005korean, yeon2013korean}.

\subsection{Chinese}
\paragraph{Word Order}: Chinese uses Subject-Verb-Object (SVO) word order. For example, "我(I)吃(eat)苹果(apple)" (I eat an apple).
\paragraph{Grammar}: Chinese uses tones to distinguish meaning between words. It lacks particles, relying on word order for grammatical structure.
\paragraph{Writing System}: Chinese characters (hanzi) are logographic, each character representing a meaning rather than a sound.
\paragraph{Differences from English}: Chinese uses tones and characters to convey meaning, whereas English relies on a phonetic alphabet and lacks tonal distinctions~\citep{norman1988chinese, ramsey1987languages}.

\subsection{Japanese}
\paragraph{Word Order}: Japanese uses Subject-Object-Verb (SOV) word order. For example, "私(I)は りんご(appe)を 食べる(eat)" (I eat an apple).
\paragraph{Grammar}: Japanese uses particles to denote grammatical roles. It has a complex system of honorifics similar to Korean.
\paragraph{Writing System}: Japanese combines phonetic scripts (hiragana and katakana) with logographic kanji.
\paragraph{Differences from English}: Japanese uses a mix of phonetic and logographic scripts and particles for grammatical relationships, unlike English which uses a purely phonetic alphabet and prepositions~\citep{shibatani1990languages, tsujimura2013introduction}.

\subsection{Key Differences from English}
\paragraph{Word Order}: English typically follows SVO structure, while Korean and Japanese use SOV, and Chinese also uses SVO.
\paragraph{Writing Systems}: English uses the Latin alphabet, a phonetic system. Korean uses Hangul, a phonetic script, while Japanese combines phonetic scripts with kanji, and Chinese uses logographic characters.
\paragraph{Grammar}: English uses prepositions and fixed word order, while Korean and Japanese use particles, and Chinese relies on word order without particles.

\section{Benchmark Tasks}\label{ap:Benchmarks}
Table~\ref{ap:dataInfo} shows detailed numbers about benchmark data.
\begin{table*}[]
\begin{adjustbox}{width=\linewidth}
\begin{tabular}{|c|ccc|cccc|}
\hline
               & \multicolumn{3}{c|}{SA}                                              & \multicolumn{4}{c|}{NLI}                                                                                 \\ \hline
benchmark name & \multicolumn{1}{c|}{GLUE SA} & \multicolumn{1}{c|}{NSMC}   & Douban  & \multicolumn{1}{c|}{GLUE-MNLI} & \multicolumn{1}{c|}{KLUE-NLI} & \multicolumn{1}{c|}{ocNLI}   & JNLI     \\ \hline
language       & \multicolumn{1}{c|}{English} & \multicolumn{1}{c|}{Korean} & Chinese & \multicolumn{1}{c|}{English}   & \multicolumn{1}{c|}{Korean}   & \multicolumn{1}{c|}{Chinese} & Japanese \\ \hline
size           & \multicolumn{1}{c|}{67,350 (67,350)}   & \multicolumn{1}{c|}{150,000 (22,000)} & 5       & \multicolumn{1}{c|}{392,703 (300,000)}    & \multicolumn{1}{c|}{27,998 (20,000)}    & \multicolumn{1}{c|}{50,000 (12,000)}   & 20,073 (12,000)        \\ \hline
\end{tabular}
\end{adjustbox}
\caption{Experiment Data Information: Total Amount (Amount Invested in the Experiment)}
\label{ap:dataInfo}
\end{table*}

\subsection{Sentiment Analysis}
Sentiment analysis, also known as opinion mining, is a field of study within natural language processing (NLP) that focuses on determining the sentiment expressed in a piece of text. This could range from binary classification (positive or negative) to more nuanced distinctions (such as very positive, positive, neutral, negative, very negative). Sentiment analysis is commonly used in various applications such as analyzing customer feedback, monitoring social media, and understanding public opinion. Techniques for sentiment analysis include machine learning approaches, lexicon-based approaches, and more recently, deep learning methods that leverage large datasets and neural network architectures.

We used the GLUE benchmark's SA~\citep{wang2018glue} for English, the NSMC~\citep{Park:2016} for Korean, and the Douban movie review for Chinese~\footnote{https://github.com/kakayuw/Sentiment-Analysis-Based-on-Douban-Movie-Critics}.

\subsection{Natural Language Inference (NLI)}
NLI is a crucial task in the field of natural language processing that involves determining the logical relationship between a pair of sentences. Given a premise and a hypothesis, the goal of NLI is to classify the relationship as entailment, contradiction, or neutral. NLI is essential for understanding and reasoning about human language, and it has applications in question answering, summarization, and dialogue systems. The development of large-scale datasets like SNLI and MNLI has significantly advanced the performance of NLI models, particularly those based on deep learning frameworks such as transformers.

We used the MNLI from the GLUE benchmark for English, the NLI from the KLUE benchmark~\citep{park2021klue} for Korean, the JNLI for Japanese~\cite{kurihara-etal-2022-jglue}, and the ocNLI for Chinese~\cite{ocnli}.

\section{Baseline detail}\label{ap:baselines}

\begin{figure}[h]
    \centering
    \includegraphics[width=\linewidth]{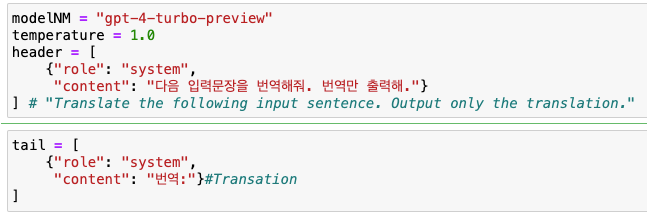}
    \caption{Information of GPT4turbo to generate NMT\textsubscript{GPT-4} data}
    \label{fig:gpt4prompt}
\end{figure}
\paragraph{GPT-4 Translated Data (NMT\textsubscript{GPT-4})}
We constructed translation data using GPT-4 turbo. Figure~\ref{fig:gpt4prompt} shows the details of the API and prompts used.

\paragraph{Local Machine Translated Data (NMT\textsubscript{mT5}*)}
 
We trained the mT5 model with translation data and used it in our experiments. Below are the details of the instructions used.
\begin{itemize}
    \item prefix = 'Translate into Korean: '
\end{itemize}

\section{Implementation Details} \label{ap:impDetail}
\subsection{Development Environment}

\subsubsection{Hardware Specifications}
\begin{itemize}
    \item \textbf{GPU}: NVIDIA A100 40GB
\end{itemize}

\subsubsection{Software Environment}
\begin{itemize}
    \item \textbf{Operating System}: Ubuntu 20.04.5
    \item \textbf{Programming Language}: Python 3.8.10
\end{itemize}

\subsubsection{Key Libraries and Tools}
\begin{table}[h!]
\centering
\begin{tabular}{|l|l|}
\hline
\textbf{Library/Tool} & \textbf{Version} \\
\hline
PyTorch       & 2.1.2  \\
Torchvision   & 0.16.2 \\
Torchaudio    & 2.1.2  \\
Transformers  & 4.38.2 \\
Accelerate    & 0.27.2 \\
OpenAI        & 0.27.7 \\
\hline
\end{tabular}
\end{table}

\section{Detailed Results}\label{ap:resultFull}
Tables~\ref{tab:sawhole} and ~\ref{tab:NLIwhole} show the whole result of SA and NLI experiments.
\begin{table*}[]
\begin{adjustbox}{width=\linewidth}
\begin{tabular}{|cccccccccccccccccccc|}
\hline
\multicolumn{1}{|c|}{datasize}   & \multicolumn{1}{c|}{1k}    & \multicolumn{1}{c|}{2k}    & \multicolumn{1}{c|}{3k}    & \multicolumn{1}{c|}{4k}    & \multicolumn{1}{c|}{5k}    & \multicolumn{1}{c|}{6k}    & \multicolumn{1}{c|}{7k}    & \multicolumn{1}{c|}{8k}    & \multicolumn{1}{c|}{9k}    & \multicolumn{1}{c|}{10k}   & \multicolumn{1}{c|}{12k}   & \multicolumn{1}{c|}{14k}   & \multicolumn{1}{c|}{15k}   & \multicolumn{1}{c|}{16k}   & \multicolumn{1}{c|}{18k}   & \multicolumn{1}{c|}{20k}   & \multicolumn{1}{c|}{30k}   & \multicolumn{1}{c|}{40k}   & 50k   \\ \hline
\multicolumn{20}{|c|}{baselines}                                                                                                                                                                                                                                                                                                                                                                                                                                                                                                                                                   \\ \hline
\multicolumn{1}{|c|}{KO}         & \multicolumn{1}{c|}{0.822} & \multicolumn{1}{c|}{0.829} & \multicolumn{1}{c|}{0.830} & \multicolumn{1}{c|}{0.816} & \multicolumn{1}{c|}{0.830} & \multicolumn{1}{c|}{0.838} & \multicolumn{1}{c|}{0.844} & \multicolumn{1}{c|}{0.849} & \multicolumn{1}{c|}{0.848} & \multicolumn{1}{c|}{0.851} & \multicolumn{1}{c|}{0.847} & \multicolumn{1}{c|}{0.849} & \multicolumn{1}{c|}{0.849} & \multicolumn{1}{c|}{0.852} & \multicolumn{1}{c|}{0.848} & \multicolumn{1}{c|}{0.857} & \multicolumn{1}{c|}{-}     & \multicolumn{1}{c|}{-}     & -     \\ \hline
\multicolumn{1}{|c|}{EN}         & \multicolumn{1}{c|}{0.689} & \multicolumn{1}{c|}{0.735} & \multicolumn{1}{c|}{0.735} & \multicolumn{1}{c|}{0.729} & \multicolumn{1}{c|}{0.742} & \multicolumn{1}{c|}{0.740} & \multicolumn{1}{c|}{0.733} & \multicolumn{1}{c|}{0.736} & \multicolumn{1}{c|}{0.742} & \multicolumn{1}{c|}{0.742} & \multicolumn{1}{c|}{0.747} & \multicolumn{1}{c|}{0.744} & \multicolumn{1}{c|}{0.747} & \multicolumn{1}{c|}{0.745} & \multicolumn{1}{c|}{0.748} & \multicolumn{1}{c|}{0.748} & \multicolumn{1}{c|}{0.749} & \multicolumn{1}{c|}{0.753} & 0.754 \\ \hline
\multicolumn{1}{|c|}{TR-GPT4}    & \multicolumn{1}{c|}{0.742} & \multicolumn{1}{c|}{0.741} & \multicolumn{1}{c|}{0.766} & \multicolumn{1}{c|}{0.738} & \multicolumn{1}{c|}{0.772} & \multicolumn{1}{c|}{0.767} & \multicolumn{1}{c|}{0.738} & \multicolumn{1}{c|}{0.772} & \multicolumn{1}{c|}{0.779} & \multicolumn{1}{c|}{0.788} & \multicolumn{1}{c|}{0.767} & \multicolumn{1}{c|}{0.769} & \multicolumn{1}{c|}{0.782} & \multicolumn{1}{c|}{0.782} & \multicolumn{1}{c|}{0.793} & \multicolumn{1}{c|}{0.787} & \multicolumn{1}{c|}{-}     & \multicolumn{1}{c|}{-}     & -     \\ \hline
\multicolumn{1}{|c|}{TR-mT5-5k}  & \multicolumn{1}{c|}{0.645} & \multicolumn{1}{c|}{0.645} & \multicolumn{1}{c|}{0.645} & \multicolumn{1}{c|}{0.645} & \multicolumn{1}{c|}{0.645} & \multicolumn{1}{c|}{0.645} & \multicolumn{1}{c|}{0.645} & \multicolumn{1}{c|}{0.644} & \multicolumn{1}{c|}{0.645} & \multicolumn{1}{c|}{0.645} & \multicolumn{1}{c|}{0.645} & \multicolumn{1}{c|}{0.645} & \multicolumn{1}{c|}{0.648} & \multicolumn{1}{c|}{0.645} & \multicolumn{1}{c|}{0.645} & \multicolumn{1}{c|}{0.645} & \multicolumn{1}{c|}{-}     & \multicolumn{1}{c|}{-}     & -     \\ \hline
\multicolumn{1}{|c|}{TR-mT5-10k} & \multicolumn{1}{c|}{0.646} & \multicolumn{1}{c|}{0.525} & \multicolumn{1}{c|}{0.491} & \multicolumn{1}{c|}{0.533} & \multicolumn{1}{c|}{0.379} & \multicolumn{1}{c|}{0.548} & \multicolumn{1}{c|}{0.302} & \multicolumn{1}{c|}{0.330} & \multicolumn{1}{c|}{0.472} & \multicolumn{1}{c|}{0.402} & \multicolumn{1}{c|}{0.349} & \multicolumn{1}{c|}{0.645} & \multicolumn{1}{c|}{0.376} & \multicolumn{1}{c|}{0.331} & \multicolumn{1}{c|}{0.277} & \multicolumn{1}{c|}{0.497} & \multicolumn{1}{c|}{-}     & \multicolumn{1}{c|}{-}     & -     \\ \hline
\multicolumn{1}{|c|}{TR-mT5-15k} & \multicolumn{1}{c|}{0.709} & \multicolumn{1}{c|}{0.727} & \multicolumn{1}{c|}{0.690} & \multicolumn{1}{c|}{0.674} & \multicolumn{1}{c|}{0.637} & \multicolumn{1}{c|}{0.648} & \multicolumn{1}{c|}{0.590} & \multicolumn{1}{c|}{0.686} & \multicolumn{1}{c|}{0.670} & \multicolumn{1}{c|}{0.700} & \multicolumn{1}{c|}{0.623} & \multicolumn{1}{c|}{0.604} & \multicolumn{1}{c|}{0.355} & \multicolumn{1}{c|}{0.248} & \multicolumn{1}{c|}{0.441} & \multicolumn{1}{c|}{0.727} & \multicolumn{1}{c|}{-}     & \multicolumn{1}{c|}{-}     & -     \\ \hline
\multicolumn{1}{|c|}{TR-mT5-20k} & \multicolumn{1}{c|}{0.702} & \multicolumn{1}{c|}{0.712} & \multicolumn{1}{c|}{0.711} & \multicolumn{1}{c|}{0.723} & \multicolumn{1}{c|}{0.680} & \multicolumn{1}{c|}{0.700} & \multicolumn{1}{c|}{0.635} & \multicolumn{1}{c|}{0.459} & \multicolumn{1}{c|}{0.612} & \multicolumn{1}{c|}{0.657} & \multicolumn{1}{c|}{0.594} & \multicolumn{1}{c|}{0.700} & \multicolumn{1}{c|}{0.681} & \multicolumn{1}{c|}{0.665} & \multicolumn{1}{c|}{0.673} & \multicolumn{1}{c|}{0.634} & \multicolumn{1}{c|}{-}     & \multicolumn{1}{c|}{-}     & -     \\ \hline
\multicolumn{20}{|c|}{LangAlign}                                                                                                                                                                                                                                                                                                                                                                                                                                                                                                                                                   \\ \hline
\multicolumn{1}{|c|}{AE5k}       & \multicolumn{1}{c|}{0.753} & \multicolumn{1}{c|}{0.755} & \multicolumn{1}{c|}{0.757} & \multicolumn{1}{c|}{0.765} & \multicolumn{1}{c|}{0.764} & \multicolumn{1}{c|}{0.770} & \multicolumn{1}{c|}{0.772} & \multicolumn{1}{c|}{0.775} & \multicolumn{1}{c|}{0.776} & \multicolumn{1}{c|}{0.780} & \multicolumn{1}{c|}{0.786} & \multicolumn{1}{c|}{0.787} & \multicolumn{1}{c|}{0.790} & \multicolumn{1}{c|}{0.791} & \multicolumn{1}{c|}{0.794} & \multicolumn{1}{c|}{0.797} & \multicolumn{1}{c|}{0.801} & \multicolumn{1}{c|}{0.802} & 0.806 \\ \hline
\multicolumn{1}{|c|}{AE10k}      & \multicolumn{1}{c|}{0.756} & \multicolumn{1}{c|}{0.742} & \multicolumn{1}{c|}{0.741} & \multicolumn{1}{c|}{0.761} & \multicolumn{1}{c|}{0.765} & \multicolumn{1}{c|}{0.759} & \multicolumn{1}{c|}{0.757} & \multicolumn{1}{c|}{0.768} & \multicolumn{1}{c|}{0.783} & \multicolumn{1}{c|}{0.773} & \multicolumn{1}{c|}{0.762} & \multicolumn{1}{c|}{0.784} & \multicolumn{1}{c|}{0.790} & \multicolumn{1}{c|}{0.790} & \multicolumn{1}{c|}{0.783} & \multicolumn{1}{c|}{0.793} & \multicolumn{1}{c|}{0.797} & \multicolumn{1}{c|}{0.790} & 0.789 \\ \hline
\multicolumn{1}{|c|}{AE15k}      & \multicolumn{1}{c|}{0.750} & \multicolumn{1}{c|}{0.767} & \multicolumn{1}{c|}{0.767} & \multicolumn{1}{c|}{0.774} & \multicolumn{1}{c|}{0.781} & \multicolumn{1}{c|}{0.776} & \multicolumn{1}{c|}{0.777} & \multicolumn{1}{c|}{0.782} & \multicolumn{1}{c|}{0.786} & \multicolumn{1}{c|}{0.790} & \multicolumn{1}{c|}{0.792} & \multicolumn{1}{c|}{0.794} & \multicolumn{1}{c|}{0.795} & \multicolumn{1}{c|}{0.796} & \multicolumn{1}{c|}{0.798} & \multicolumn{1}{c|}{0.803} & \multicolumn{1}{c|}{0.811} & \multicolumn{1}{c|}{0.811} & 0.820 \\ \hline
\multicolumn{1}{|c|}{AE20k}      & \multicolumn{1}{c|}{0.771} & \multicolumn{1}{c|}{0.767} & \multicolumn{1}{c|}{0.762} & \multicolumn{1}{c|}{0.786} & \multicolumn{1}{c|}{0.788} & \multicolumn{1}{c|}{0.773} & \multicolumn{1}{c|}{0.759} & \multicolumn{1}{c|}{0.781} & \multicolumn{1}{c|}{0.804} & \multicolumn{1}{c|}{0.804} & \multicolumn{1}{c|}{0.796} & \multicolumn{1}{c|}{0.788} & \multicolumn{1}{c|}{0.802} & \multicolumn{1}{c|}{0.809} & \multicolumn{1}{c|}{0.802} & \multicolumn{1}{c|}{0.812} & \multicolumn{1}{c|}{0.808} & \multicolumn{1}{c|}{0.813} & 0.822 \\ \hline
\multicolumn{1}{|c|}{FC5k}       & \multicolumn{1}{c|}{0.731} & \multicolumn{1}{c|}{0.740} & \multicolumn{1}{c|}{0.755} & \multicolumn{1}{c|}{0.750} & \multicolumn{1}{c|}{0.744} & \multicolumn{1}{c|}{0.749} & \multicolumn{1}{c|}{0.748} & \multicolumn{1}{c|}{0.755} & \multicolumn{1}{c|}{0.766} & \multicolumn{1}{c|}{0.763} & \multicolumn{1}{c|}{0.769} & \multicolumn{1}{c|}{0.773} & \multicolumn{1}{c|}{0.775} & \multicolumn{1}{c|}{0.777} & \multicolumn{1}{c|}{0.775} & \multicolumn{1}{c|}{0.779} & \multicolumn{1}{c|}{0.777} & \multicolumn{1}{c|}{0.789} & 0.784 \\ \hline
\multicolumn{1}{|c|}{FC10k}      & \multicolumn{1}{c|}{0.733} & \multicolumn{1}{c|}{0.750} & \multicolumn{1}{c|}{0.764} & \multicolumn{1}{c|}{0.733} & \multicolumn{1}{c|}{0.739} & \multicolumn{1}{c|}{0.763} & \multicolumn{1}{c|}{0.738} & \multicolumn{1}{c|}{0.745} & \multicolumn{1}{c|}{0.778} & \multicolumn{1}{c|}{0.754} & \multicolumn{1}{c|}{0.758} & \multicolumn{1}{c|}{0.766} & \multicolumn{1}{c|}{0.764} & \multicolumn{1}{c|}{0.774} & \multicolumn{1}{c|}{0.786} & \multicolumn{1}{c|}{0.772} & \multicolumn{1}{c|}{0.776} & \multicolumn{1}{c|}{0.785} & 0.790 \\ \hline
\multicolumn{1}{|c|}{FC15k}      & \multicolumn{1}{c|}{0.735} & \multicolumn{1}{c|}{0.747} & \multicolumn{1}{c|}{0.747} & \multicolumn{1}{c|}{0.754} & \multicolumn{1}{c|}{0.766} & \multicolumn{1}{c|}{0.763} & \multicolumn{1}{c|}{0.767} & \multicolumn{1}{c|}{0.763} & \multicolumn{1}{c|}{0.780} & \multicolumn{1}{c|}{0.768} & \multicolumn{1}{c|}{0.768} & \multicolumn{1}{c|}{0.783} & \multicolumn{1}{c|}{0.782} & \multicolumn{1}{c|}{0.790} & \multicolumn{1}{c|}{0.788} & \multicolumn{1}{c|}{0.795} & \multicolumn{1}{c|}{0.801} & \multicolumn{1}{c|}{0.813} & 0.810 \\ \hline
\multicolumn{1}{|c|}{FC20k}      & \multicolumn{1}{c|}{0.759} & \multicolumn{1}{c|}{0.763} & \multicolumn{1}{c|}{0.763} & \multicolumn{1}{c|}{0.760} & \multicolumn{1}{c|}{0.767} & \multicolumn{1}{c|}{0.765} & \multicolumn{1}{c|}{0.771} & \multicolumn{1}{c|}{0.773} & \multicolumn{1}{c|}{0.773} & \multicolumn{1}{c|}{0.775} & \multicolumn{1}{c|}{0.776} & \multicolumn{1}{c|}{0.787} & \multicolumn{1}{c|}{0.789} & \multicolumn{1}{c|}{0.778} & \multicolumn{1}{c|}{0.789} & \multicolumn{1}{c|}{0.800} & \multicolumn{1}{c|}{0.794} & \multicolumn{1}{c|}{0.797} & 0.819 \\ \hline
\multicolumn{20}{|c|}{Ablation -   LangAlign Omission}                                                                                                                                                                                                                                                                                                                                                                                                                                                                                                                             \\ \hline
\multicolumn{1}{|c|}{5k}         & \multicolumn{1}{c|}{0.700} & \multicolumn{1}{c|}{0.692} & \multicolumn{1}{c|}{0.713} & \multicolumn{1}{c|}{0.693} & \multicolumn{1}{c|}{0.708} & \multicolumn{1}{c|}{0.709} & \multicolumn{1}{c|}{0.698} & \multicolumn{1}{c|}{0.729} & \multicolumn{1}{c|}{0.719} & \multicolumn{1}{c|}{0.715} & \multicolumn{1}{c|}{0.710} & \multicolumn{1}{c|}{0.719} & \multicolumn{1}{c|}{0.727} & \multicolumn{1}{c|}{0.731} & \multicolumn{1}{c|}{0.744} & \multicolumn{1}{c|}{0.728} & \multicolumn{1}{c|}{0.717} & \multicolumn{1}{c|}{0.735} & 0.745 \\ \hline
\multicolumn{1}{|c|}{10k}        & \multicolumn{1}{c|}{0.697} & \multicolumn{1}{c|}{0.691} & \multicolumn{1}{c|}{0.702} & \multicolumn{1}{c|}{0.682} & \multicolumn{1}{c|}{0.721} & \multicolumn{1}{c|}{0.721} & \multicolumn{1}{c|}{0.690} & \multicolumn{1}{c|}{0.702} & \multicolumn{1}{c|}{0.734} & \multicolumn{1}{c|}{0.699} & \multicolumn{1}{c|}{0.706} & \multicolumn{1}{c|}{0.709} & \multicolumn{1}{c|}{0.720} & \multicolumn{1}{c|}{0.739} & \multicolumn{1}{c|}{0.739} & \multicolumn{1}{c|}{0.734} & \multicolumn{1}{c|}{0.737} & \multicolumn{1}{c|}{0.729} & 0.733 \\ \hline
\multicolumn{1}{|c|}{15k}        & \multicolumn{1}{c|}{0.686} & \multicolumn{1}{c|}{0.708} & \multicolumn{1}{c|}{0.712} & \multicolumn{1}{c|}{0.700} & \multicolumn{1}{c|}{0.712} & \multicolumn{1}{c|}{0.711} & \multicolumn{1}{c|}{0.709} & \multicolumn{1}{c|}{0.717} & \multicolumn{1}{c|}{0.737} & \multicolumn{1}{c|}{0.736} & \multicolumn{1}{c|}{0.738} & \multicolumn{1}{c|}{0.728} & \multicolumn{1}{c|}{0.732} & \multicolumn{1}{c|}{0.729} & \multicolumn{1}{c|}{0.732} & \multicolumn{1}{c|}{0.742} & \multicolumn{1}{c|}{0.749} & \multicolumn{1}{c|}{0.736} & 0.744 \\ \hline
\multicolumn{1}{|c|}{20k}        & \multicolumn{1}{c|}{0.701} & \multicolumn{1}{c|}{0.712} & \multicolumn{1}{c|}{0.715} & \multicolumn{1}{c|}{0.698} & \multicolumn{1}{c|}{0.725} & \multicolumn{1}{c|}{0.726} & \multicolumn{1}{c|}{0.711} & \multicolumn{1}{c|}{0.725} & \multicolumn{1}{c|}{0.721} & \multicolumn{1}{c|}{0.724} & \multicolumn{1}{c|}{0.729} & \multicolumn{1}{c|}{0.740} & \multicolumn{1}{c|}{0.737} & \multicolumn{1}{c|}{0.739} & \multicolumn{1}{c|}{0.740} & \multicolumn{1}{c|}{0.737} & \multicolumn{1}{c|}{0.759} & \multicolumn{1}{c|}{0.748} & 0.752 \\ \hline
\multicolumn{20}{|c|}{Ablation -   Datashare}                                                                                                                                                                                                                                                                                                                                                                                                                                                                                                                                      \\ \hline
\multicolumn{1}{|c|}{AE5k}       & \multicolumn{1}{c|}{0.756} & \multicolumn{1}{c|}{0.755} & \multicolumn{1}{c|}{0.769} & \multicolumn{1}{c|}{0.755} & \multicolumn{1}{c|}{0.783} & \multicolumn{1}{c|}{0.776} & \multicolumn{1}{c|}{0.760} & \multicolumn{1}{c|}{0.785} & \multicolumn{1}{c|}{0.786} & \multicolumn{1}{c|}{0.799} & \multicolumn{1}{c|}{0.785} & \multicolumn{1}{c|}{0.785} & \multicolumn{1}{c|}{0.783} & \multicolumn{1}{c|}{0.795} & \multicolumn{1}{c|}{0.804} & \multicolumn{1}{c|}{0.801} & \multicolumn{1}{c|}{0.805} & \multicolumn{1}{c|}{0.806} & 0.809 \\ \hline
\multicolumn{1}{|c|}{AE10k}      & \multicolumn{1}{c|}{0.756} & \multicolumn{1}{c|}{0.751} & \multicolumn{1}{c|}{0.751} & \multicolumn{1}{c|}{0.754} & \multicolumn{1}{c|}{0.784} & \multicolumn{1}{c|}{0.779} & \multicolumn{1}{c|}{0.757} & \multicolumn{1}{c|}{0.777} & \multicolumn{1}{c|}{0.788} & \multicolumn{1}{c|}{0.781} & \multicolumn{1}{c|}{0.768} & \multicolumn{1}{c|}{0.787} & \multicolumn{1}{c|}{0.785} & \multicolumn{1}{c|}{0.794} & \multicolumn{1}{c|}{0.803} & \multicolumn{1}{c|}{0.800} & \multicolumn{1}{c|}{0.800} & \multicolumn{1}{c|}{0.785} & 0.793 \\ \hline
\multicolumn{1}{|c|}{AE15k}      & \multicolumn{1}{c|}{0.753} & \multicolumn{1}{c|}{0.764} & \multicolumn{1}{c|}{0.774} & \multicolumn{1}{c|}{0.752} & \multicolumn{1}{c|}{0.784} & \multicolumn{1}{c|}{0.779} & \multicolumn{1}{c|}{0.769} & \multicolumn{1}{c|}{0.786} & \multicolumn{1}{c|}{0.790} & \multicolumn{1}{c|}{0.798} & \multicolumn{1}{c|}{0.784} & \multicolumn{1}{c|}{0.792} & \multicolumn{1}{c|}{0.792} & \multicolumn{1}{c|}{0.797} & \multicolumn{1}{c|}{0.802} & \multicolumn{1}{c|}{0.802} & \multicolumn{1}{c|}{0.800} & \multicolumn{1}{c|}{0.791} & 0.800 \\ \hline
\multicolumn{1}{|c|}{AE20k}      & \multicolumn{1}{c|}{0.755} & \multicolumn{1}{c|}{0.755} & \multicolumn{1}{c|}{0.778} & \multicolumn{1}{c|}{0.783} & \multicolumn{1}{c|}{0.790} & \multicolumn{1}{c|}{0.778} & \multicolumn{1}{c|}{0.757} & \multicolumn{1}{c|}{0.784} & \multicolumn{1}{c|}{0.799} & \multicolumn{1}{c|}{0.805} & \multicolumn{1}{c|}{0.789} & \multicolumn{1}{c|}{0.783} & \multicolumn{1}{c|}{0.789} & \multicolumn{1}{c|}{0.809} & \multicolumn{1}{c|}{0.806} & \multicolumn{1}{c|}{0.821} & \multicolumn{1}{c|}{0.814} & \multicolumn{1}{c|}{0.822} & 0.814 \\ \hline
\multicolumn{1}{|c|}{FC5k}       & \multicolumn{1}{c|}{0.752} & \multicolumn{1}{c|}{0.745} & \multicolumn{1}{c|}{0.776} & \multicolumn{1}{c|}{0.751} & \multicolumn{1}{c|}{0.749} & \multicolumn{1}{c|}{0.776} & \multicolumn{1}{c|}{0.752} & \multicolumn{1}{c|}{0.760} & \multicolumn{1}{c|}{0.791} & \multicolumn{1}{c|}{0.793} & \multicolumn{1}{c|}{0.775} & \multicolumn{1}{c|}{0.778} & \multicolumn{1}{c|}{0.781} & \multicolumn{1}{c|}{0.783} & \multicolumn{1}{c|}{0.780} & \multicolumn{1}{c|}{0.794} & \multicolumn{1}{c|}{0.794} & \multicolumn{1}{c|}{0.812} & 0.810 \\ \hline
\multicolumn{1}{|c|}{FC10k}      & \multicolumn{1}{c|}{0.738} & \multicolumn{1}{c|}{0.754} & \multicolumn{1}{c|}{0.774} & \multicolumn{1}{c|}{0.741} & \multicolumn{1}{c|}{0.764} & \multicolumn{1}{c|}{0.772} & \multicolumn{1}{c|}{0.746} & \multicolumn{1}{c|}{0.758} & \multicolumn{1}{c|}{0.791} & \multicolumn{1}{c|}{0.767} & \multicolumn{1}{c|}{0.770} & \multicolumn{1}{c|}{0.772} & \multicolumn{1}{c|}{0.782} & \multicolumn{1}{c|}{0.782} & \multicolumn{1}{c|}{0.789} & \multicolumn{1}{c|}{0.779} & \multicolumn{1}{c|}{0.780} & \multicolumn{1}{c|}{0.788} & 0.792 \\ \hline
\multicolumn{1}{|c|}{FC15k}      & \multicolumn{1}{c|}{0.750} & \multicolumn{1}{c|}{0.750} & \multicolumn{1}{c|}{0.766} & \multicolumn{1}{c|}{0.756} & \multicolumn{1}{c|}{0.779} & \multicolumn{1}{c|}{0.780} & \multicolumn{1}{c|}{0.757} & \multicolumn{1}{c|}{0.781} & \multicolumn{1}{c|}{0.790} & \multicolumn{1}{c|}{0.775} & \multicolumn{1}{c|}{0.777} & \multicolumn{1}{c|}{0.786} & \multicolumn{1}{c|}{0.789} & \multicolumn{1}{c|}{0.795} & \multicolumn{1}{c|}{0.803} & \multicolumn{1}{c|}{0.798} & \multicolumn{1}{c|}{0.782} & \multicolumn{1}{c|}{0.798} & 0.803 \\ \hline
\multicolumn{1}{|c|}{FC20k}      & \multicolumn{1}{c|}{0.757} & \multicolumn{1}{c|}{0.755} & \multicolumn{1}{c|}{0.769} & \multicolumn{1}{c|}{0.761} & \multicolumn{1}{c|}{0.783} & \multicolumn{1}{c|}{0.773} & \multicolumn{1}{c|}{0.756} & \multicolumn{1}{c|}{0.783} & \multicolumn{1}{c|}{0.782} & \multicolumn{1}{c|}{0.787} & \multicolumn{1}{c|}{0.779} & \multicolumn{1}{c|}{0.786} & \multicolumn{1}{c|}{0.786} & \multicolumn{1}{c|}{0.783} & \multicolumn{1}{c|}{0.802} & \multicolumn{1}{c|}{0.801} & \multicolumn{1}{c|}{0.803} & \multicolumn{1}{c|}{0.791} & 0.792 \\ \hline
\end{tabular}
\end{adjustbox}
\caption{Whole Result of Sentiment Analysis Experiment}
\label{tab:sawhole}
\end{table*}
\begin{table*}[]
\begin{adjustbox}{width=\linewidth}
\begin{tabular}{|cccccccccccccccccccc|}
\hline
\multicolumn{1}{|c|}{datasize}   & \multicolumn{1}{c|}{1k}    & \multicolumn{1}{c|}{2k}    & \multicolumn{1}{c|}{3k}    & \multicolumn{1}{c|}{4k}    & \multicolumn{1}{c|}{5k}    & \multicolumn{1}{c|}{6k}    & \multicolumn{1}{c|}{7k}    & \multicolumn{1}{c|}{8k}    & \multicolumn{1}{c|}{9k}    & \multicolumn{1}{c|}{10k}   & \multicolumn{1}{c|}{12k}   & \multicolumn{1}{c|}{14k}   & \multicolumn{1}{c|}{15k}   & \multicolumn{1}{c|}{16k}   & \multicolumn{1}{c|}{18k}   & \multicolumn{1}{c|}{20k}   & \multicolumn{1}{c|}{30k}   & \multicolumn{1}{c|}{40k}   & 50k   \\ \hline
\multicolumn{20}{|c|}{baselines}                                                                                                                                                                                                                                                                                                                                                                                                                                                                                                                                                   \\ \hline
\multicolumn{1}{|c|}{KO}         & \multicolumn{1}{c|}{0.718} & \multicolumn{1}{c|}{0.722} & \multicolumn{1}{c|}{0.740} & \multicolumn{1}{c|}{0.763} & \multicolumn{1}{c|}{0.739} & \multicolumn{1}{c|}{0.757} & \multicolumn{1}{c|}{0.761} & \multicolumn{1}{c|}{0.747} & \multicolumn{1}{c|}{0.773} & \multicolumn{1}{c|}{0.792} & \multicolumn{1}{c|}{0.783} & \multicolumn{1}{c|}{0.800} & \multicolumn{1}{c|}{0.795} & \multicolumn{1}{c|}{0.775} & \multicolumn{1}{c|}{0.789} & \multicolumn{1}{c|}{0.812} & \multicolumn{1}{c|}{-}     & \multicolumn{1}{c|}{-}     & -     \\ \hline
\multicolumn{1}{|c|}{EN}         & \multicolumn{1}{c|}{0.591} & \multicolumn{1}{c|}{0.649} & \multicolumn{1}{c|}{0.655} & \multicolumn{1}{c|}{0.671} & \multicolumn{1}{c|}{0.688} & \multicolumn{1}{c|}{0.675} & \multicolumn{1}{c|}{0.685} & \multicolumn{1}{c|}{0.702} & \multicolumn{1}{c|}{0.697} & \multicolumn{1}{c|}{0.709} & \multicolumn{1}{c|}{0.710} & \multicolumn{1}{c|}{0.709} & \multicolumn{1}{c|}{0.713} & \multicolumn{1}{c|}{0.717} & \multicolumn{1}{c|}{0.719} & \multicolumn{1}{c|}{0.724} & \multicolumn{1}{c|}{0.723} & \multicolumn{1}{c|}{0.724} & 0.731 \\ \hline
\multicolumn{1}{|c|}{TR-GPT4}    & \multicolumn{1}{c|}{0.650} & \multicolumn{1}{c|}{0.655} & \multicolumn{1}{c|}{0.719} & \multicolumn{1}{c|}{0.727} & \multicolumn{1}{c|}{0.726} & \multicolumn{1}{c|}{0.762} & \multicolumn{1}{c|}{0.757} & \multicolumn{1}{c|}{0.755} & \multicolumn{1}{c|}{0.665} & \multicolumn{1}{c|}{0.785} & \multicolumn{1}{c|}{0.773} & \multicolumn{1}{c|}{0.779} & \multicolumn{1}{c|}{0.766} & \multicolumn{1}{c|}{0.766} & \multicolumn{1}{c|}{0.767} & \multicolumn{1}{c|}{0.774} & \multicolumn{1}{c|}{-}     & \multicolumn{1}{c|}{-}     & -     \\ \hline
\multicolumn{1}{|c|}{TR-mT5-5k}  & \multicolumn{1}{c|}{0.341} & \multicolumn{1}{c|}{0.341} & \multicolumn{1}{c|}{0.341} & \multicolumn{1}{c|}{0.341} & \multicolumn{1}{c|}{0.341} & \multicolumn{1}{c|}{0.341} & \multicolumn{1}{c|}{0.341} & \multicolumn{1}{c|}{0.341} & \multicolumn{1}{c|}{0.341} & \multicolumn{1}{c|}{0.341} & \multicolumn{1}{c|}{0.341} & \multicolumn{1}{c|}{0.341} & \multicolumn{1}{c|}{0.341} & \multicolumn{1}{c|}{0.341} & \multicolumn{1}{c|}{0.341} & \multicolumn{1}{c|}{0.341} & \multicolumn{1}{c|}{-}     & \multicolumn{1}{c|}{-}     & -     \\ \hline
\multicolumn{1}{|c|}{TR-mT5-10k} & \multicolumn{1}{c|}{0.541} & \multicolumn{1}{c|}{0.424} & \multicolumn{1}{c|}{0.412} & \multicolumn{1}{c|}{0.482} & \multicolumn{1}{c|}{0.302} & \multicolumn{1}{c|}{0.453} & \multicolumn{1}{c|}{0.299} & \multicolumn{1}{c|}{0.246} & \multicolumn{1}{c|}{0.340} & \multicolumn{1}{c|}{0.389} & \multicolumn{1}{c|}{0.282} & \multicolumn{1}{c|}{0.605} & \multicolumn{1}{c|}{0.475} & \multicolumn{1}{c|}{0.267} & \multicolumn{1}{c|}{0.237} & \multicolumn{1}{c|}{0.468} & \multicolumn{1}{c|}{-}     & \multicolumn{1}{c|}{-}     & -     \\ \hline
\multicolumn{1}{|c|}{TR-mT5-15k} & \multicolumn{1}{c|}{0.619} & \multicolumn{1}{c|}{0.628} & \multicolumn{1}{c|}{0.628} & \multicolumn{1}{c|}{0.645} & \multicolumn{1}{c|}{0.550} & \multicolumn{1}{c|}{0.627} & \multicolumn{1}{c|}{0.565} & \multicolumn{1}{c|}{0.625} & \multicolumn{1}{c|}{0.557} & \multicolumn{1}{c|}{0.680} & \multicolumn{1}{c|}{0.574} & \multicolumn{1}{c|}{0.594} & \multicolumn{1}{c|}{0.289} & \multicolumn{1}{c|}{0.188} & \multicolumn{1}{c|}{0.405} & \multicolumn{1}{c|}{0.690} & \multicolumn{1}{c|}{-}     & \multicolumn{1}{c|}{-}     & -     \\ \hline
\multicolumn{1}{|c|}{TR-mT5-20k} & \multicolumn{1}{c|}{0.603} & \multicolumn{1}{c|}{0.604} & \multicolumn{1}{c|}{0.651} & \multicolumn{1}{c|}{0.679} & \multicolumn{1}{c|}{0.614} & \multicolumn{1}{c|}{0.664} & \multicolumn{1}{c|}{0.617} & \multicolumn{1}{c|}{0.352} & \multicolumn{1}{c|}{0.473} & \multicolumn{1}{c|}{0.598} & \multicolumn{1}{c|}{0.540} & \multicolumn{1}{c|}{0.662} & \multicolumn{1}{c|}{0.650} & \multicolumn{1}{c|}{0.608} & \multicolumn{1}{c|}{0.624} & \multicolumn{1}{c|}{0.579} & \multicolumn{1}{c|}{-}     & \multicolumn{1}{c|}{-}     & -     \\ \hline
\multicolumn{20}{|c|}{LangAlign}                                                                                                                                                                                                                                                                                                                                                                                                                                                                                                                                                   \\ \hline
\multicolumn{1}{|c|}{AE5k}       & \multicolumn{1}{c|}{0.654} & \multicolumn{1}{c|}{0.665} & \multicolumn{1}{c|}{0.708} & \multicolumn{1}{c|}{0.735} & \multicolumn{1}{c|}{0.714} & \multicolumn{1}{c|}{0.692} & \multicolumn{1}{c|}{0.705} & \multicolumn{1}{c|}{0.712} & \multicolumn{1}{c|}{0.616} & \multicolumn{1}{c|}{0.764} & \multicolumn{1}{c|}{0.758} & \multicolumn{1}{c|}{0.801} & \multicolumn{1}{c|}{0.718} & \multicolumn{1}{c|}{0.747} & \multicolumn{1}{c|}{0.742} & \multicolumn{1}{c|}{0.783} & \multicolumn{1}{c|}{0.766} & \multicolumn{1}{c|}{0.788} & 0.777 \\ \hline
\multicolumn{1}{|c|}{AE10k}      & \multicolumn{1}{c|}{0.667} & \multicolumn{1}{c|}{0.659} & \multicolumn{1}{c|}{0.655} & \multicolumn{1}{c|}{0.710} & \multicolumn{1}{c|}{0.702} & \multicolumn{1}{c|}{0.713} & \multicolumn{1}{c|}{0.740} & \multicolumn{1}{c|}{0.693} & \multicolumn{1}{c|}{0.717} & \multicolumn{1}{c|}{0.712} & \multicolumn{1}{c|}{0.762} & \multicolumn{1}{c|}{0.781} & \multicolumn{1}{c|}{0.753} & \multicolumn{1}{c|}{0.753} & \multicolumn{1}{c|}{0.740} & \multicolumn{1}{c|}{0.753} & \multicolumn{1}{c|}{0.770} & \multicolumn{1}{c|}{0.749} & 0.773 \\ \hline
\multicolumn{1}{|c|}{AE15k}      & \multicolumn{1}{c|}{0.655} & \multicolumn{1}{c|}{0.654} & \multicolumn{1}{c|}{0.682} & \multicolumn{1}{c|}{0.729} & \multicolumn{1}{c|}{0.700} & \multicolumn{1}{c|}{0.760} & \multicolumn{1}{c|}{0.709} & \multicolumn{1}{c|}{0.754} & \multicolumn{1}{c|}{0.684} & \multicolumn{1}{c|}{0.775} & \multicolumn{1}{c|}{0.745} & \multicolumn{1}{c|}{0.755} & \multicolumn{1}{c|}{0.744} & \multicolumn{1}{c|}{0.749} & \multicolumn{1}{c|}{0.756} & \multicolumn{1}{c|}{0.756} & \multicolumn{1}{c|}{0.799} & \multicolumn{1}{c|}{0.799} & 0.801 \\ \hline
\multicolumn{1}{|c|}{AE20k}      & \multicolumn{1}{c|}{0.676} & \multicolumn{1}{c|}{0.679} & \multicolumn{1}{c|}{0.707} & \multicolumn{1}{c|}{0.747} & \multicolumn{1}{c|}{0.717} & \multicolumn{1}{c|}{0.729} & \multicolumn{1}{c|}{0.727} & \multicolumn{1}{c|}{0.724} & \multicolumn{1}{c|}{0.670} & \multicolumn{1}{c|}{0.797} & \multicolumn{1}{c|}{0.751} & \multicolumn{1}{c|}{0.758} & \multicolumn{1}{c|}{0.775} & \multicolumn{1}{c|}{0.754} & \multicolumn{1}{c|}{0.763} & \multicolumn{1}{c|}{0.787} & \multicolumn{1}{c|}{0.783} & \multicolumn{1}{c|}{0.794} & 0.810 \\ \hline
\multicolumn{1}{|c|}{FC5k}       & \multicolumn{1}{c|}{0.639} & \multicolumn{1}{c|}{0.644} & \multicolumn{1}{c|}{0.705} & \multicolumn{1}{c|}{0.725} & \multicolumn{1}{c|}{0.666} & \multicolumn{1}{c|}{0.706} & \multicolumn{1}{c|}{0.714} & \multicolumn{1}{c|}{0.690} & \multicolumn{1}{c|}{0.657} & \multicolumn{1}{c|}{0.737} & \multicolumn{1}{c|}{0.729} & \multicolumn{1}{c|}{0.755} & \multicolumn{1}{c|}{0.719} & \multicolumn{1}{c|}{0.720} & \multicolumn{1}{c|}{0.732} & \multicolumn{1}{c|}{0.736} & \multicolumn{1}{c|}{0.764} & \multicolumn{1}{c|}{0.754} & 0.776 \\ \hline
\multicolumn{1}{|c|}{FC10k}      & \multicolumn{1}{c|}{0.627} & \multicolumn{1}{c|}{0.653} & \multicolumn{1}{c|}{0.695} & \multicolumn{1}{c|}{0.684} & \multicolumn{1}{c|}{0.675} & \multicolumn{1}{c|}{0.697} & \multicolumn{1}{c|}{0.697} & \multicolumn{1}{c|}{0.721} & \multicolumn{1}{c|}{0.664} & \multicolumn{1}{c|}{0.730} & \multicolumn{1}{c|}{0.752} & \multicolumn{1}{c|}{0.738} & \multicolumn{1}{c|}{0.762} & \multicolumn{1}{c|}{0.718} & \multicolumn{1}{c|}{0.734} & \multicolumn{1}{c|}{0.741} & \multicolumn{1}{c|}{0.757} & \multicolumn{1}{c|}{0.761} & 0.787 \\ \hline
\multicolumn{1}{|c|}{FC15k}      & \multicolumn{1}{c|}{0.643} & \multicolumn{1}{c|}{0.665} & \multicolumn{1}{c|}{0.695} & \multicolumn{1}{c|}{0.735} & \multicolumn{1}{c|}{0.680} & \multicolumn{1}{c|}{0.754} & \multicolumn{1}{c|}{0.746} & \multicolumn{1}{c|}{0.735} & \multicolumn{1}{c|}{0.650} & \multicolumn{1}{c|}{0.710} & \multicolumn{1}{c|}{0.721} & \multicolumn{1}{c|}{0.743} & \multicolumn{1}{c|}{0.776} & \multicolumn{1}{c|}{0.736} & \multicolumn{1}{c|}{0.743} & \multicolumn{1}{c|}{0.762} & \multicolumn{1}{c|}{0.781} & \multicolumn{1}{c|}{0.778} & 0.781 \\ \hline
\multicolumn{1}{|c|}{FC20k}      & \multicolumn{1}{c|}{0.662} & \multicolumn{1}{c|}{0.657} & \multicolumn{1}{c|}{0.680} & \multicolumn{1}{c|}{0.746} & \multicolumn{1}{c|}{0.710} & \multicolumn{1}{c|}{0.700} & \multicolumn{1}{c|}{0.772} & \multicolumn{1}{c|}{0.698} & \multicolumn{1}{c|}{0.751} & \multicolumn{1}{c|}{0.762} & \multicolumn{1}{c|}{0.747} & \multicolumn{1}{c|}{0.787} & \multicolumn{1}{c|}{0.771} & \multicolumn{1}{c|}{0.737} & \multicolumn{1}{c|}{0.740} & \multicolumn{1}{c|}{0.764} & \multicolumn{1}{c|}{0.765} & \multicolumn{1}{c|}{0.748} & 0.774 \\ \hline
\multicolumn{20}{|c|}{Ablation -   LangAlign Omission}                                                                                                                                                                                                                                                                                                                                                                                                                                                                                                                             \\ \hline
\multicolumn{1}{|c|}{5k}         & \multicolumn{1}{c|}{0.601} & \multicolumn{1}{c|}{0.598} & \multicolumn{1}{c|}{0.659} & \multicolumn{1}{c|}{0.652} & \multicolumn{1}{c|}{0.641} & \multicolumn{1}{c|}{0.671} & \multicolumn{1}{c|}{0.710} & \multicolumn{1}{c|}{0.630} & \multicolumn{1}{c|}{0.649} & \multicolumn{1}{c|}{0.697} & \multicolumn{1}{c|}{0.663} & \multicolumn{1}{c|}{0.675} & \multicolumn{1}{c|}{0.667} & \multicolumn{1}{c|}{0.651} & \multicolumn{1}{c|}{0.681} & \multicolumn{1}{c|}{0.686} & \multicolumn{1}{c|}{0.683} & \multicolumn{1}{c|}{0.716} & 0.728 \\ \hline
\multicolumn{1}{|c|}{10k}        & \multicolumn{1}{c|}{0.595} & \multicolumn{1}{c|}{0.597} & \multicolumn{1}{c|}{0.646} & \multicolumn{1}{c|}{0.628} & \multicolumn{1}{c|}{0.617} & \multicolumn{1}{c|}{0.696} & \multicolumn{1}{c|}{0.635} & \multicolumn{1}{c|}{0.597} & \multicolumn{1}{c|}{0.642} & \multicolumn{1}{c|}{0.647} & \multicolumn{1}{c|}{0.652} & \multicolumn{1}{c|}{0.681} & \multicolumn{1}{c|}{0.707} & \multicolumn{1}{c|}{0.655} & \multicolumn{1}{c|}{0.687} & \multicolumn{1}{c|}{0.701} & \multicolumn{1}{c|}{0.699} & \multicolumn{1}{c|}{0.692} & 0.714 \\ \hline
\multicolumn{1}{|c|}{15k}        & \multicolumn{1}{c|}{0.575} & \multicolumn{1}{c|}{0.600} & \multicolumn{1}{c|}{0.620} & \multicolumn{1}{c|}{0.632} & \multicolumn{1}{c|}{0.654} & \multicolumn{1}{c|}{0.670} & \multicolumn{1}{c|}{0.673} & \multicolumn{1}{c|}{0.617} & \multicolumn{1}{c|}{0.603} & \multicolumn{1}{c|}{0.707} & \multicolumn{1}{c|}{0.707} & \multicolumn{1}{c|}{0.704} & \multicolumn{1}{c|}{0.692} & \multicolumn{1}{c|}{0.660} & \multicolumn{1}{c|}{0.685} & \multicolumn{1}{c|}{0.696} & \multicolumn{1}{c|}{0.714} & \multicolumn{1}{c|}{0.713} & 0.729 \\ \hline
\multicolumn{1}{|c|}{20k}        & \multicolumn{1}{c|}{0.603} & \multicolumn{1}{c|}{0.622} & \multicolumn{1}{c|}{0.656} & \multicolumn{1}{c|}{0.688} & \multicolumn{1}{c|}{0.646} & \multicolumn{1}{c|}{0.683} & \multicolumn{1}{c|}{0.623} & \multicolumn{1}{c|}{0.673} & \multicolumn{1}{c|}{0.647} & \multicolumn{1}{c|}{0.695} & \multicolumn{1}{c|}{0.697} & \multicolumn{1}{c|}{0.719} & \multicolumn{1}{c|}{0.695} & \multicolumn{1}{c|}{0.684} & \multicolumn{1}{c|}{0.691} & \multicolumn{1}{c|}{0.706} & \multicolumn{1}{c|}{0.722} & \multicolumn{1}{c|}{0.726} & 0.717 \\ \hline
\multicolumn{20}{|c|}{Ablation -   Datashare}                                                                                                                                                                                                                                                                                                                                                                                                                                                                                                                                      \\ \hline
\multicolumn{1}{|c|}{AE5k}       & \multicolumn{1}{c|}{0.649} & \multicolumn{1}{c|}{0.671} & \multicolumn{1}{c|}{0.724} & \multicolumn{1}{c|}{0.726} & \multicolumn{1}{c|}{0.721} & \multicolumn{1}{c|}{0.711} & \multicolumn{1}{c|}{0.726} & \multicolumn{1}{c|}{0.715} & \multicolumn{1}{c|}{0.661} & \multicolumn{1}{c|}{0.797} & \multicolumn{1}{c|}{0.759} & \multicolumn{1}{c|}{0.780} & \multicolumn{1}{c|}{0.721} & \multicolumn{1}{c|}{0.765} & \multicolumn{1}{c|}{0.753} & \multicolumn{1}{c|}{0.763} & \multicolumn{1}{c|}{0.789} & \multicolumn{1}{c|}{0.788} & 0.800 \\ \hline
\multicolumn{1}{|c|}{AE10k}      & \multicolumn{1}{c|}{0.666} & \multicolumn{1}{c|}{0.640} & \multicolumn{1}{c|}{0.692} & \multicolumn{1}{c|}{0.749} & \multicolumn{1}{c|}{0.707} & \multicolumn{1}{c|}{0.753} & \multicolumn{1}{c|}{0.681} & \multicolumn{1}{c|}{0.743} & \multicolumn{1}{c|}{0.701} & \multicolumn{1}{c|}{0.727} & \multicolumn{1}{c|}{0.728} & \multicolumn{1}{c|}{0.754} & \multicolumn{1}{c|}{0.747} & \multicolumn{1}{c|}{0.774} & \multicolumn{1}{c|}{0.770} & \multicolumn{1}{c|}{0.764} & \multicolumn{1}{c|}{0.775} & \multicolumn{1}{c|}{0.776} & 0.782 \\ \hline
\multicolumn{1}{|c|}{AE15k}      & \multicolumn{1}{c|}{0.652} & \multicolumn{1}{c|}{0.673} & \multicolumn{1}{c|}{0.715} & \multicolumn{1}{c|}{0.745} & \multicolumn{1}{c|}{0.702} & \multicolumn{1}{c|}{0.731} & \multicolumn{1}{c|}{0.722} & \multicolumn{1}{c|}{0.731} & \multicolumn{1}{c|}{0.707} & \multicolumn{1}{c|}{0.748} & \multicolumn{1}{c|}{0.733} & \multicolumn{1}{c|}{0.798} & \multicolumn{1}{c|}{0.753} & \multicolumn{1}{c|}{0.745} & \multicolumn{1}{c|}{0.754} & \multicolumn{1}{c|}{0.767} & \multicolumn{1}{c|}{0.787} & \multicolumn{1}{c|}{0.763} & 0.792 \\ \hline
\multicolumn{1}{|c|}{AE20k}      & \multicolumn{1}{c|}{0.653} & \multicolumn{1}{c|}{0.676} & \multicolumn{1}{c|}{0.688} & \multicolumn{1}{c|}{0.764} & \multicolumn{1}{c|}{0.706} & \multicolumn{1}{c|}{0.713} & \multicolumn{1}{c|}{0.694} & \multicolumn{1}{c|}{0.760} & \multicolumn{1}{c|}{0.733} & \multicolumn{1}{c|}{0.802} & \multicolumn{1}{c|}{0.776} & \multicolumn{1}{c|}{0.777} & \multicolumn{1}{c|}{0.768} & \multicolumn{1}{c|}{0.766} & \multicolumn{1}{c|}{0.751} & \multicolumn{1}{c|}{0.782} & \multicolumn{1}{c|}{0.786} & \multicolumn{1}{c|}{0.807} & 0.790 \\ \hline
\multicolumn{1}{|c|}{FC5k}       & \multicolumn{1}{c|}{0.661} & \multicolumn{1}{c|}{0.650} & \multicolumn{1}{c|}{0.714} & \multicolumn{1}{c|}{0.722} & \multicolumn{1}{c|}{0.694} & \multicolumn{1}{c|}{0.711} & \multicolumn{1}{c|}{0.694} & \multicolumn{1}{c|}{0.732} & \multicolumn{1}{c|}{0.730} & \multicolumn{1}{c|}{0.762} & \multicolumn{1}{c|}{0.714} & \multicolumn{1}{c|}{0.759} & \multicolumn{1}{c|}{0.761} & \multicolumn{1}{c|}{0.714} & \multicolumn{1}{c|}{0.732} & \multicolumn{1}{c|}{0.767} & \multicolumn{1}{c|}{0.763} & \multicolumn{1}{c|}{0.800} & 0.789 \\ \hline
\multicolumn{1}{|c|}{FC10k}      & \multicolumn{1}{c|}{0.638} & \multicolumn{1}{c|}{0.670} & \multicolumn{1}{c|}{0.718} & \multicolumn{1}{c|}{0.710} & \multicolumn{1}{c|}{0.695} & \multicolumn{1}{c|}{0.765} & \multicolumn{1}{c|}{0.739} & \multicolumn{1}{c|}{0.728} & \multicolumn{1}{c|}{0.735} & \multicolumn{1}{c|}{0.744} & \multicolumn{1}{c|}{0.776} & \multicolumn{1}{c|}{0.785} & \multicolumn{1}{c|}{0.755} & \multicolumn{1}{c|}{0.736} & \multicolumn{1}{c|}{0.749} & \multicolumn{1}{c|}{0.737} & \multicolumn{1}{c|}{0.743} & \multicolumn{1}{c|}{0.767} & 0.785 \\ \hline
\multicolumn{1}{|c|}{FC15k}      & \multicolumn{1}{c|}{0.646} & \multicolumn{1}{c|}{0.655} & \multicolumn{1}{c|}{0.707} & \multicolumn{1}{c|}{0.704} & \multicolumn{1}{c|}{0.723} & \multicolumn{1}{c|}{0.729} & \multicolumn{1}{c|}{0.760} & \multicolumn{1}{c|}{0.686} & \multicolumn{1}{c|}{0.664} & \multicolumn{1}{c|}{0.776} & \multicolumn{1}{c|}{0.744} & \multicolumn{1}{c|}{0.748} & \multicolumn{1}{c|}{0.744} & \multicolumn{1}{c|}{0.721} & \multicolumn{1}{c|}{0.763} & \multicolumn{1}{c|}{0.781} & \multicolumn{1}{c|}{0.749} & \multicolumn{1}{c|}{0.789} & 0.775 \\ \hline
\multicolumn{1}{|c|}{FC20k}      & \multicolumn{1}{c|}{0.657} & \multicolumn{1}{c|}{0.649} & \multicolumn{1}{c|}{0.704} & \multicolumn{1}{c|}{0.710} & \multicolumn{1}{c|}{0.707} & \multicolumn{1}{c|}{0.720} & \multicolumn{1}{c|}{0.724} & \multicolumn{1}{c|}{0.748} & \multicolumn{1}{c|}{0.727} & \multicolumn{1}{c|}{0.752} & \multicolumn{1}{c|}{0.777} & \multicolumn{1}{c|}{0.741} & \multicolumn{1}{c|}{0.750} & \multicolumn{1}{c|}{0.751} & \multicolumn{1}{c|}{0.763} & \multicolumn{1}{c|}{0.772} & \multicolumn{1}{c|}{0.776} & \multicolumn{1}{c|}{0.778} & 0.773 \\ \hline
\end{tabular}
\end{adjustbox}
\caption{Whole Result of NLI Experiment}
\label{tab:NLIwhole}
\end{table*}

\subsection{Detailed Cosine}\label{ap:cosineDetail}
Table~\ref{tab:cosineFull} shows the whole result of Cosine similarity experiments.
\begin{table}[]
\centering
\label{tab:combined}
\begin{adjustbox}{width=\linewidth}
\begin{tabular}{ccccc}
\hline
\multicolumn{1}{l|}{}       & 5k    & 10k   & 15k   & 20k   \\ \hline
\multicolumn{1}{l|}{GPT-EN} & 0.879 & 0.868 & 0.849 & 0.887 \\
\multicolumn{1}{l|}{EN-AE}  & 0.885 & 0.881 & 0.87  & 0.903 \\
\multicolumn{1}{l|}{EN-FC}  & 0.887 & 0.878 & 0.872 & 0.903 \\
\multicolumn{1}{l|}{GPT-AE} & 0.952 & 0.953 & 0.948 & 0.951 \\
\multicolumn{1}{l|}{GPT-FC} & 0.953 & 0.942 & 0.952 & 0.958 \\ \hline
\multicolumn{5}{c}{SA cosine similarity}                    \\ \hline
\multicolumn{1}{l|}{}       & 5k    & 10k   & 15k   & 20k   \\ \hline
\multicolumn{1}{l|}{GPT-EN} & 0.889 & 0.831 & 0.869 & 0.882 \\ 
\multicolumn{1}{l|}{EN-AE}  & 0.899 & 0.836 & 0.887 & 0.894 \\
\multicolumn{1}{l|}{EN-FC}  & 0.883 & 0.839 & 0.884 & 0.895 \\
\multicolumn{1}{l|}{GPT-AE} & 0.975 & 0.92  & 0.955 & 0.961 \\
\multicolumn{1}{l|}{GPT-FC} & 0.969 & 0.928 & 0.954 & 0.967 \\ \hline
\multicolumn{5}{c}{NLI cosine similarity}                  
\end{tabular}
\end{adjustbox}
\caption{Measurement results of cosine similarity between data overall}
\label{tab:cosineFull}
\end{table}
\subsection{\revModelNM} \label{ap:revFull}
Table~\ref{tab:revFull} shows the whole result of \revModelNM experiments.
\begin{table}[]
\begin{adjustbox}{width=\linewidth}
\begin{tabular}{l|ll|lll}
\hline
           & \multicolumn{2}{c|}{SA} & \multicolumn{3}{c}{NLI} \\ \cline{2-6} 
           & 20k         & 50k       & 20k    & 50k    & 300k  \\ \hline
EN         & 0.748       & 0.754     & 0.724  & 0.731  & 0.761 \\
TR-GPT     & 0.787       & -         & 0.774  & -      & -     \\
Rev-AE 5k  & 0.7489      & 0.743     & 0.7335 & 0.7318 & 0.751 \\
Rev-AE 10k & 0.7597      & 0.757     & 0.7188 & 0.7639 & 0.768 \\
Rev-AE 15k & 0.7581      & 0.786     & 0.7475 & 0.7546 & 0.798 \\
Rev-AE 20k & 0.766       & 0.797     & 0.7619 & 0.7661 & 0.812 \\
Rev-FC 5k  & 0.7446      & 0.734     & 0.7029 & 0.7493 & 0.737 \\
Rev-FC 10k & 0.7584      & 0.734     & 0.7191 & 0.7501 & 0.739 \\
Rev-FC 15k & 0.7552      & 0.772     & 0.7245 & 0.7552 & 0.776 \\
Rev-FC 20k & 0.7619      & 0.777     & 0.7489 & 0.752  & 0.819 \\ \hline
\end{tabular}
\end{adjustbox}
\caption{\revModelNM Experiment Result}
\label{tab:revFull}
\end{table}

\begin{table}[]
\begin{adjustbox}{width=\linewidth}
\begin{tabular}{c|cccc|cccc}
\hline
              & \multicolumn{4}{c|}{SA}       & \multicolumn{4}{c}{NLI}       \\ \hline
datasize      & 5K    & 10K   & 15K   & 20K   & 5K    & 10K   & 15K   & 20K   \\ \hline
LoRA-TA       & 0.746 & 0.759 & 0.759 & 0.753 & 0.714 & 0.733 & 0.741 & 0.738 \\
Unsup. SimSCE & 0.614 & 0.541 & 0.465 & 0.597 & 0.688 & 0.701 & 0.699 & 0.700 \\
NMT\_GPT4     & 0.772 & 0.788 & 0.782 & 0.787 & 0.726 & 0.785 & 0.766 & 0.774 \\
LA-AE-15K     & 0.781 & 0.790 & 0.795 & 0.803 & 0.700 & 0.775 & 0.744 & 0.756 \\
LA-FC-15K     & 0.766 & 0.768 & 0.782 & 0.795 & 0.680 & 0.710 & 0.776 & 0.762 \\ \hline
\end{tabular}
\end{adjustbox}
\caption{Experimental Results using LoRA-TA and SimSCE}\label{tab:loraFull}
\end{table}
\section{Comparison with Other Architectures}\label{ap:otherArchitectures}
In this study, we analyzed the performance improvement efficiency of \modelName relative to its data construction costs and compared it with various data construction methods. Additionally, we evaluated \modelName against other studies that, while not directly related to data construction, share conceptual similarities in task learning and processing performance.

\subsection{LoRA-based Task Adaptation (LoRA-TA)}
LoRA-based task adaptation is a technique that keeps all parameters of a large model frozen and fine-tunes the model with a small number of additional parameters by learning only low-rank matrices. We evaluated the performance of the LoRA-TA model fine-tuned using NMT\textsubscript{GPT-4} data.

The LoRA configuration consisted of a rank \(r = 8\), scaling factor \(\alpha = 8\), and a dropout probability of 0.1, applied to the query and key modules of the attention layer. This allowed the model to reduce memory usage and computational complexity by learning only low-rank matrices while keeping the original parameters fixed. Training was performed with a batch size of 16, a learning rate of 1e-5, and a maximum sequence length of 256, using the AdamW optimizer to achieve efficient task adaptation.

\subsection{Unsupervised-SimSCE (SimSCE)}
SimSCE is a method that enhances embedding representations using unlabeled data. We applied SimSCE to the XLM-RoBERTa model using NMT\textsubscript{GPT-4} data, followed by task tuning.

SimCSE embedding training was conducted with a batch size of 16, a learning rate of 1e-5, a maximum sequence length of 256, and for 10 epochs using a contrastive learning loss function with a temperature parameter of 0.05. During training, the same sentence was encoded twice, and noise was added through dropout to perform contrastive learning. This process allowed the model to learn to keep embeddings of the same sentence closer together while pushing embeddings of different sentences further apart, thereby strengthening sentence representation.

\subsection{Experimental Results}
Table~\ref{tab:loraFull} shows the benchmark performance of models with LoRA-TA and SimSCE applied.  
The experimental results indicate that LoRA-TA showed slightly lower performance compared to \modelName or models trained directly on Korean data. This is likely due to the XLM-RoBERTa-base model having the capacity to train all parameters, which reveals the limitations of LoRA’s approach that only trains a subset of parameters.  
The model with SimSCE also did not show performance improvement. This aligns with the concept of our study, which assumes a scenario with limited available training data. Under such conditions, it is challenging to have enough data to effectively train embeddings, suggesting that SimSCE’s potential for enhancing representation could not be fully realized.

\end{document}